%% file: iclr2026_conference.tex
\documentclass{article} %
\usepackage{iclr2026_conference,times}
\iclrfinalcopy

\input{math_commands.tex}

\usepackage{seqsplit}

\usepackage{hyperref}
\usepackage{url}
\usepackage{chngcntr}
\counterwithout{table}{section} %
\usepackage{graphicx}
\usepackage{algorithm}
\usepackage{algpseudocode}
\usepackage{listings}
\usepackage{xcolor}
\usepackage{amssymb}
\usepackage{booktabs}
\usepackage{multirow}
\usepackage[table]{xcolor} %
\usepackage{makecell} %
\lstdefinestyle{timesplain}{
  language={},                    %
  basicstyle=\footnotesize\rmfamily, %
  showstringspaces=false,
  breaklines=true,
  columns=fullflexible,
  keepspaces=true,
  frame=single,
  rulecolor=\color{black!25},
  backgroundcolor=\color{black!3},
  numbers=none,
  xleftmargin=0.5em,
  framexleftmargin=0.5em,
  aboveskip=0.8\baselineskip,
  belowskip=0.8\baselineskip,
  captionpos=b
}
\usepackage{tcolorbox,enumitem,ragged2e}
\tcbset{
  promptbox/.style={
    colback=black!2, colframe=black!20, boxrule=0.3pt,
    arc=1mm, left=6pt, right=6pt, top=6pt, bottom=6pt}
}

\usepackage{caption}
\usepackage{enumitem}
\setitemize{noitemsep,topsep=0pt,parsep=0pt,partopsep=0pt}

\usepackage{graphicx}
\title{UniVid: The Open-Source Unified Video Model}

\author{\textbf{Jiabin Luo$^{1*}$~
Junhui Lin$^{2*}$~
Zeyu Zhang$^{1*\dag}$~
Biao Wu$^{3*}$}~
\textbf{
Meng Fang$^{3}$~
Ling Chen$^{3}$~
Hao Tang$^{1\ddag}$}\\[0.4em]
$^{1}$Peking University~~
$^{2}$AI Geeks~~
$^{3}$Australian Artificial Intelligence Institute\\[0.3em]
\footnotesize $^*$Equal contribution.
$^\dag$Project lead.
$^\ddag$Corresponding author: bjdxtanghao@gmail.com.\\[0.3em]
}

\begin{document}

\maketitle

\begin{figure}[h]
\centering
\includegraphics[width=\columnwidth]{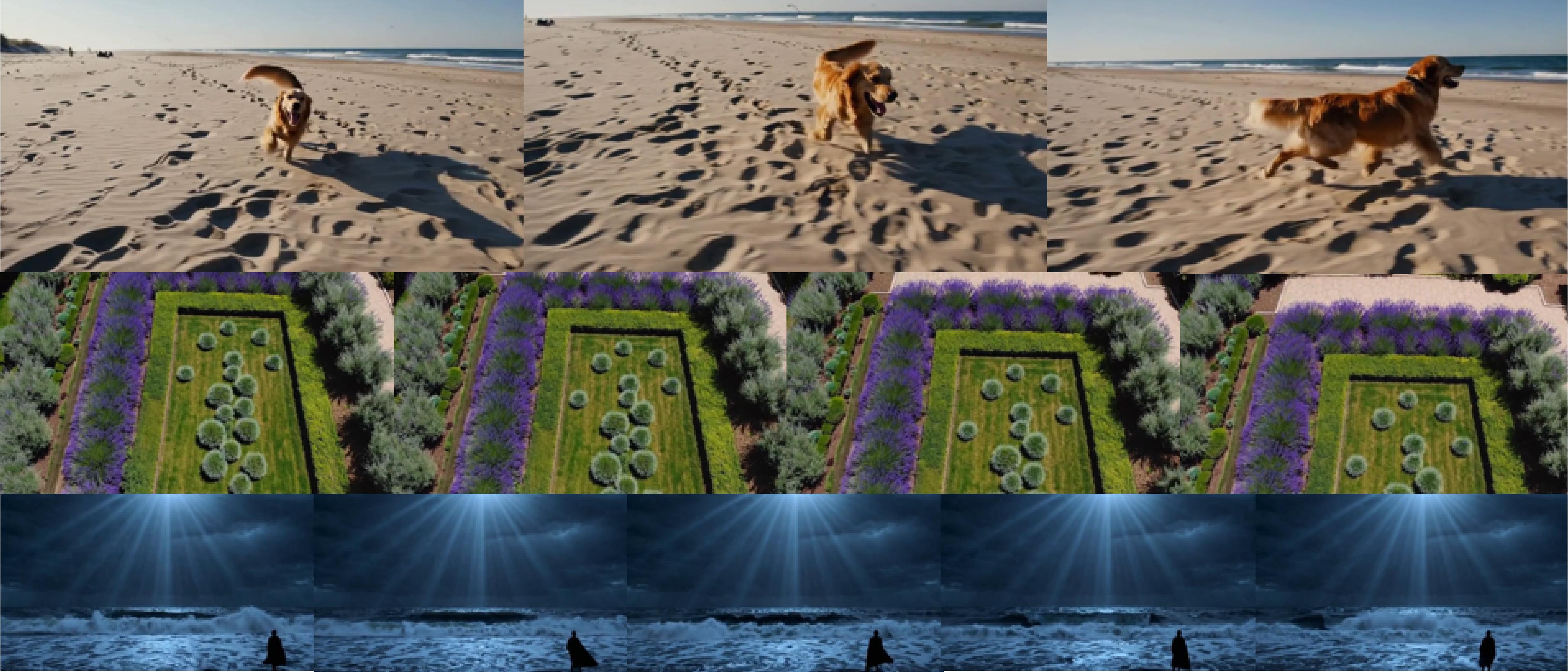}
\caption{We present \textbf{UniVid}, an open-source unified video model for both understanding and generation tasks. Our model requires only a small amount of high-quality data for fine-tuning, achieveing competitive results across various tasks.}
\label{fig:teaser}
\end{figure}

\begin{abstract}

Unified video modeling combining generation and understanding capabilities is increasingly important, yet faces two key challenges: maintaining semantic faithfulness during flow-based generation due to text-visual token imbalance and the suboptimality of uniform cross-modal attention across the flow trajectory, and efficiently extending image-centric MLLMs to video without costly retraining. We present \textbf{UniVid}, a unified architecture that couples an MLLM with a diffusion decoder through a lightweight adapter, enabling both video understanding and generation. We introduce \textit{Temperature Modality Alignment} to improve prompt adherence and \textit{Pyramid Reflection} for efficient temporal reasoning via dynamic keyframe selection. Extensive experiments on standard benchmarks demonstrate the state-of-the-art performance of our unified video model, achieving a \textbf{2.2\%} improvement on VBench-Long total score compared to the previous SOTA method EasyAnimateV5.1, and \textbf{1.0\%} and \textbf{3.3\%} accuracy gains on MSVD-QA and ActivityNet-QA, respectively, compared with the best prior 7B baselines. 
Code: \url{https://github.com/AIGeeksGroup/UniVid}.
Website: \url{https://aigeeksgroup.github.io/UniVid}.

\end{abstract}

\section{Introduction}
Video intelligence encompasses two core capabilities: generation and understanding. Generation enables content creation, simulation, and data augmentation through diffusion and flow models~\citep{shi2020improvingimagecaptioningbetter,SDXL,Wan,SVD}. Understanding powers perception, retrieval, analytics, and human-computer interaction via multimodal LLMs~\citep{Qwen2-VL,InternVL,Video-LLava,Qwen2.5-VL}. Real-world applications increasingly demand unified systems that combine both capabilities within a single framework. Recent efforts toward unified video modeling have converged on two paradigms. The first is an autoregressive (AR)–centric route: all modalities (text, images, video) are projected into a shared discrete token space and a single Transformer is trained with next-token prediction over multimodal sequences; representative examples include Emu3~\citep{Emu3} and Chameleon~\citep{Chameleon}. The second is a hybrid diffusion–AR route: a multimodal AR backbone governs understanding and control signals, while a diffusion video decoder renders high-fidelity frames from high-level visual tokens; recent works such as Transfusion~\citep{transfusion} and Show-O~\citep{showo} follow this pattern. In this work, we adopt the hybrid route to retain high-quality rendering while leveraging an MLLM for semantic control and interpretability.

However, even within this hybrid setting, unified video modeling faces two key challenges. First, maintaining semantically faithful conditioning in video diffusion across the flow trajectory is difficult. Text prompts convey high-level intent but under-specify pixel-aligned details; in MM-DiT-style ~\cite{MM-DiT} models, the cross-modal signal can be diluted by the numerical imbalance between few text tokens and many visual tokens, and the role of guidance is inherently timestep-dependent—early steps benefit more from strong semantic constraints, whereas later steps benefit from visual detail refinement, yielding prompt–video drift that worsens with longer, higher-resolution clips. Second, extending image-centric MLLMs to video faces two key challenges: the computational cost of temporal modeling (dedicated encoders, long-context handling, large-scale training) that risks destabilizing existing capabilities, and the mismatch between video's vast temporal information and the typically small subset relevant to any question. Traditional approaches either process all frames uniformly, causing inefficiency and noise, or use fixed sampling that may miss critical evidence. Furthermore, different question types demand different strategies—static questions need distinctive keyframes while dynamic questions require understanding temporal transitions. 

To address these challenges, our motivation is twofold. First, on the generation side, we leverage multimodal understanding to construct structure-aware tokens in the language space that encode both global semantics and localized cues; these tokens are used as faithful semantic conditioning for a diffusion video decoder, and we schedule cross-modal attention over flow steps so that early integration emphasizes textual intent while later steps emphasize visual refinement. Second, on the understanding side, we develop an adaptive evidence selection approach that extends image-centric MLLMs to video without substantial architectural changes. This requires a mechanism that can iteratively explore and refine the evidence set based on feedback, balance exploration of new frames with exploitation of current evidence, and learn from failure signals to improve future selections. This suggests a sequential decision-making framework, but rather than traditional parameter updates, we implement a form of verbal test-time reinforcement learning. We develop Pyramid Reflection, where policy improvement occurs through natural language refinement—the Reflector verbally adjusts search queries based on feedback, while SigLIP2 ~\citep{SigLip2} enables query-driven keyframe selection that iteratively expands or prunes the evidence set.

Hence, we propose \textbf{UniVid}, a unified architecture that couples a multimodal LLM with a diffusion video decoder via a lightweight conditioning adapter: the LLM ingests text and salient visual evidence and outputs rich semantic understandable tokens that both support reasoning and condition the decoder for text/image-to-video generation. To stabilize guidance in MM-DiT ~\citep{MM-DiT}, we introduce \textit{Temperature Modality Alignment}, a timestep-aware, temperature-adjusted cross-modal attention schedule that emphasizes semantic intent early and visual refinement late, mitigating text suppression and improving prompt faithfulness. To enable efficient understanding with minimal change, we introduce \textit{Pyramid Reflection}, which implements sequential decision-making through SigLIP2-based keyframe selection and an Actor–Evaluator–Reflector loop that verbally adjusts search strategies while progressively expanding or pruning context.
Through extensive evaluation on standard benchmarks, we validate the superior capability of our unified approach, which consistently outperforms existing methods across multiple video-centric tasks, demonstrating the potential of unified modeling for comprehensive video intelligence.

Our contribution can be summarized below: 
\begin{itemize}[leftmargin=*]
  \item We introduce \textbf{UniVid}, a unified paradigm that couples an MLLM with a diffusion video decoder via a lightweight conditioning adapter; the MLLM produces rich, understandable semantic tokens that both support reasoning and condition text/image-to-video generation.
  \item We propose \textit{Temperature Modality Alignment}, a timestep-aware, temperature-adjusted cross-modal attention schedule in MM-DiT that strengthens early semantic guidance and later shifts emphasis to visual refinement; we further develop \textit{Pyramid Reflection} with SigLIP2-based keyframe selection to enable efficient temporal reasoning with minimal architectural change and training.
  \item We conduct comprehensive experiments on MSVD-QA~\citep{msrvtt}, MSRVTT-QA~\citep{msrvtt}, TGIF-QA~\citep{TGIF}, and ActivityNet-QA~\citep{ActivityNet-QA} for understanding, and on VBench for generation, demonstrating competitive performance and efficiency. Ablations verify the contribution of each component.

\end{itemize}

\begin{figure}[t] 
  \centering
  \includegraphics[width=\textwidth]{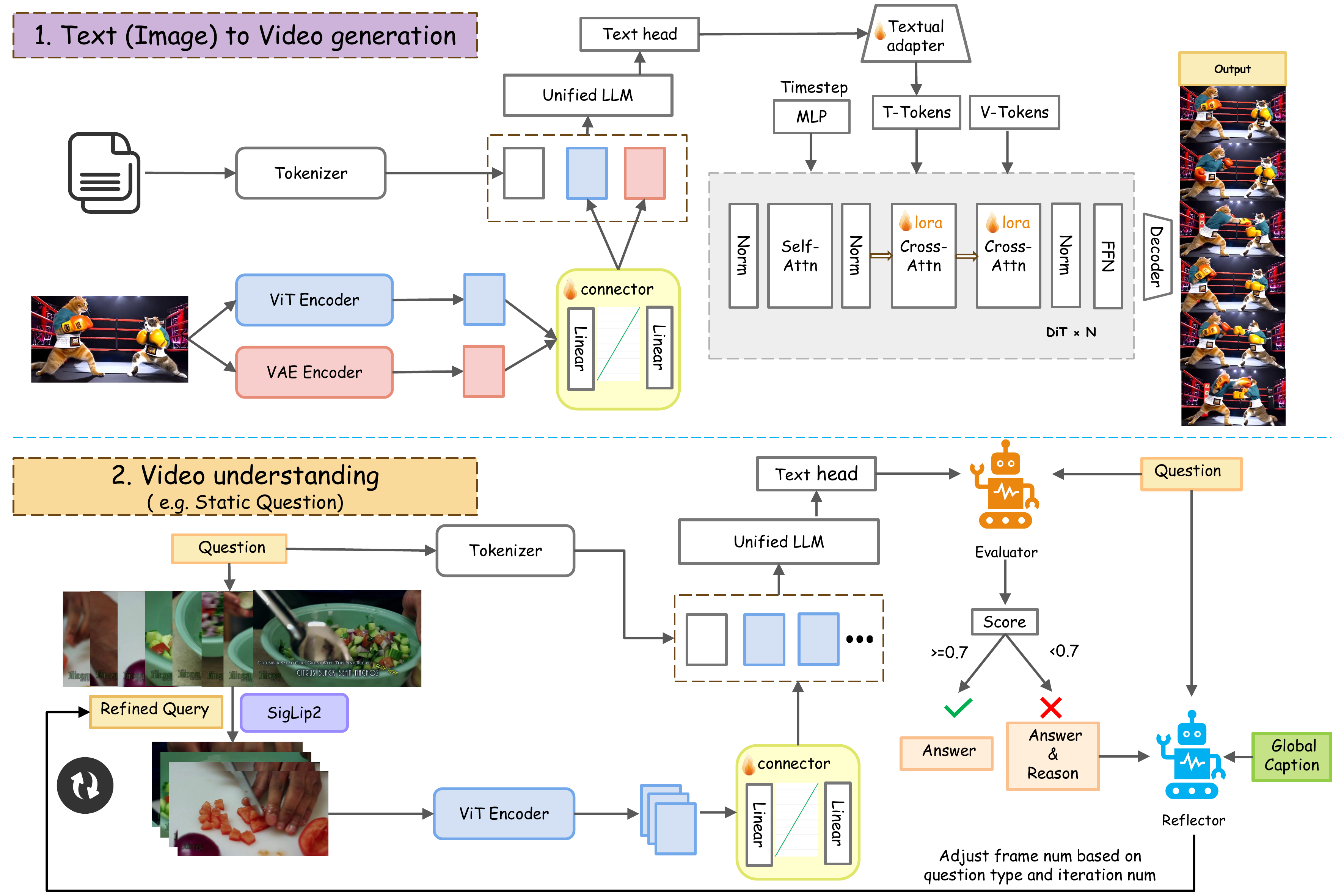}
  \caption{Overall architecture of our proposed UniVid for unified video understanding and generation. UniVid couples an autoregressive-based MLLM with a DiT-based diffusion decoder. The MLLM's outputs are linked through a lightweight adapter to interface with the Wan~\citep{Wan} backbone, forming the generation branch, while simultaneously passing through the Pyramid Reflection module to connect with the LLM, thereby establishing the understanding branch. } 
  \label{fig:architecture}
  \vspace{-0.4cm}
\end{figure}

\section{Related Work}

\paragraph{Video generation.}
Video generation has seen remarkable advancements with the rise of diffusion models and generative adversarial networks tailored for temporal data. Recent diffusion or flow based frameworks, such as Video Diffusion Models ~\citep{vdm}, Imagen Video ~\citep{Imagen}, VideoCrafter2 ~\citep{videocrafter} and Stable Video Diffusion ~\citep{blattmann2023stable}, have produced high-fidelity clips with improved temporal consistency, enabling applications in creative generation and simulation~\citep{liu2025fpsattention,shi2025presentagent}. Latent diffusion techniques ~\citep{blattmann2023align} further improve efficiency by operating in compressed latent spaces, enabling scalable video generation. In parallel, GAN methods like MoCoGAN ~\citep{MoCoGAN} and StyleGAN-V ~\citep{stylegan-v} explore alternative formulations. Despite these advances, maintaining long-term temporal consistency in extended sequences remains challenging, as summarized by recent surveys and analyses ~\citep{vdm_survey,consistency_survey}. %

\paragraph{Video understanding.}
Recent progress in video understanding has been driven by transformer-based architectures and self-supervised learning paradigms that effectively model spatio-temporal relationships. Methods like MViT ~\citep{MViT}, Video Swin Transformer ~\citep{liu2022video}, TimeSformer ~\citep{timesformer} and ViViT ~\citep{vivit} have advanced the field by capturing long-range dependencies across video frames, achieving strong performance on datasets such as Kinetics-700 ~\citep{carreira2019short}. Beyond supervised training, self-supervised approaches—including masked modeling (VideoMAE ~\citep{videomae}, MaskFeat ~\citep{maskfeat}, OmniMAE ~\citep{girdhar2023omnimae}) and early contrastive methods (VideoMoCo ~\citep{pan2021videomoco})—leverage unlabeled videos to learn robust, transferable representations, reducing dependence on costly annotations and benefiting action recognition and video segmentation.

\paragraph{Unified multimodal models.}

Unified multimodal modeling has progressed from joint vision–language pretraining to architectures that support both understanding and generation across modalities. Foundational systems like CLIP ~\citep{radford2021learning} establish large-scale alignment, while BEiT-3 ~\citep{wang2023image} and UnifiedMLLM ~\citep{li2024unifiedmllm} broaden task coverage. Pushing toward unified generation, Show-o ~\citep{showo} integrates autoregression with discrete diffusion within a single Transformer to support VQA, text-to-image, and various editing tasks. In a complementary direction focused on robustness rather than general any-to-any generation, FLUID ~\citep{fluid} uses token-level distillation for cross-modal fusion. Open generalist systems then aim to unify understanding and generation end-to-end: BAGEL ~\citep{BAGEL} offers an open, decoder-only framework with parallel language and diffusion branches trained jointly, achieving competitive results across image-centric tasks, and BLIP3-o ~\citep{BILP3-o} releases a fully open family where a diffusion transformer is coupled to strong multimodal understanding, yielding unified image understanding and generation. Extending unification from images to video, Omni-Video ~\citep{Omni-Video} teaches an MLLM to emit continuous visual tokens that are adapted and consumed by a diffusion video decoder, enabling generation, editing, and understanding in one pipeline.

\section{The Proposed Method}
 
\subsection{Overview}

Our goal is a unified multimodal video model that supports both generation and understanding within a single framework. To this end, we adopt a three-stage hierarchical training recipe that first aligns the conditioning between the MLLM and the generator, then finetunes the MLLM and introduces Pyramid Reflection, which augments the understanding branch with temporal cues, and finally co-adapts both branches end-to-end. Fig.~\ref{fig:architecture} presents the overall UniVid architecture. 

\subsection{Architecture}

\noindent\textbf{Multimodal archiecture.}
The multimodal large language model serves as the core reasoning engine. Text inputs are processed through a standard tokenizer, while visual inputs follow different encoding paths depending on the target branch. For the generation branch, images are encoded using both ViT ~\citep{vit} for semantic features and VAE ~\citep{VAE} for pixel-level details. For the understanding branch, only ViT encoding is employed, as video understanding tasks primarily rely on high-level semantic understanding rather than fine-grained pixel details. The encoded visual features are then projected into the textual token space and concatenated with text tokens, allowing the LLM to output unified multimodal representations.

\noindent\textbf{Generation branch.}
The generation pathway employs a DiT-based model Wan~2.2 ~\citep{Wan} conditioned on rich semantic representations extracted from MLLM outputs through a lightweight adapter. The system processes video generation in latent space using a 3D VAE ~\citep{3DVAE}, with conditioning signals integrated via cross-attention mechanisms.

\noindent\textbf{Understanding branch.}
For video understanding, multi-frame evidence is encoded by the ViT ~\citep{vit} and fused with text; the LLM produces an initial textual answer. We then apply Pyramid Reflection, a query-driven, hierarchical loop that iteratively expands or prunes keyframe context via SigLIP2 ~\citep{SigLip2} selection and refines the frame space via an Actor–Evaluator–Reflector process, yielding the final answer without modifying the backbone. 

Conclusively, our generation builds on the MLLM’s strong comprehension, while video understanding uses Pyramid Reflection to leverage the MLLM and collaborate with an LLM for efficient and accurate answers.

\subsection{Conditional Generation with Temperature Modality Alignment}
\begin{algorithm}[!t]
\caption{Pyramid Reflection as Test-time RL}
\label{Pyramid_Reflection_algorithm}
\begin{algorithmic}[1]
\Require video $V$, question $q$
\State Uniformly sample $N{=}64$ frames; \emph{encode once and cache} visual embeddings
\State From 16 frames, summarize into a global caption $C_g$
\State Initialize state $s_1 \!\leftarrow\! (q, C_g, W{=}\varnothing)$, policy $\pi$ with mode router \textbf{expand}/\textbf{shrink}
\For{$r = 1$ to $R \le 3$}
  \State \textbf{Action}: $a_r \sim \pi(s_r)$ 
  \Statex \quad\textbf{expand}: add frames most relevant to current search text
  \Statex \quad\textbf{shrink}: prune to diverse key frames using cached similarities
  \State Update working set $W$ accordingly using cached embeddings \hfill{\small (index-only change)}
  \State \textbf{Actor}: answer using ordered $W$ conditioned on $C_g$
  \State \textbf{Evaluator}: score $\hat{r}_r \in [0,1]$ as confidence signal
  \If{$\hat{r}_r \ge \tau$} \Return answer
  \Else \textbf{Reflector}: refine the search text $q \!\leftarrow\!$ short declarative cue
  \State Update state $s_{r+1} \!\leftarrow\! (q, C_g, W)$ \hfill{\small (verbal policy improvement)}
  \EndIf
\EndFor
\State \Return fallback answer from $C_g$
\end{algorithmic}
\end{algorithm}
Given fused tokens from the understanding path, the MLLM output $Z_u$ is mapped to time-indexed conditions by a lightweight adapter $g_{\phi}$:
\begin{equation}
C_t = g_{\phi}(Z_u, t)\in\mathbb{R}^{M_t\times d_c},
\end{equation}
where $M_t$ is the number of conditioning tokens at timestep $t$ and $d_c$ is the conditioning dimension.

Let the 3D VAE define the latent trajectory $\{z_t\}$ along the flow, where $z_t \in \mathbb{R}^{H \times W \times F \times C}$ represents the latent representation with spatial dimensions $H \times W$, temporal frames $F$, and channels $C$.
The Wan~2.2 DiT predicts the velocity field under cross-attention to $C_t$, then we integrate the probability–flow ODE to obtain $\hat z_0$, which the VAE decoder converts to video frames.

Inspired by TACA~\citep{lv2025rethinkingcrossmodalinteractionmultimodal}, we adapt its finding that text is suppressed in MM-DiT ~\citep{MM-DiT} because (i) the softmax over a much larger pool of visual tokens ($N_{\text{vis}}\gg N_{\text{txt}}$) dilutes attention mass on text keys, and (ii) conditioning plays different roles across timesteps (early semantics, late detail).
We therefore strengthen the visual-to-text path in Wan~2.2 ~\citep{Wan} with a simple schedule:
\begin{equation}
\tilde S_{v\to t}(u)\;=\;\alpha_{\text{txt}}(u)\,S_{v\to t},\qquad u\in[0,1],
\end{equation}
where $u$ is the normalized flow matching progress (0 early, 1 late), $S_{v\to t}$ denotes the visual-to-text attention scores, and $\tilde S_{v\to t}(u)$ represents the modulated attention scores. The modulation factor is defined as:
\begin{equation}
\label{factor1}
\alpha_{\text{txt}}(u) = 
\begin{cases}
1 + \dfrac{\lambda_{\text{txt}}}{2}\!\left(1 + \cos\!\left(\dfrac{\pi u}{0.4}\right)\right), & u \in [0,\, 0.4], \\[6pt]
1, & u \in (0.4,\, 1],
\end{cases}
\qquad \lambda_{\text{txt}} = 0.3.
\end{equation}
Thus, text guidance is strongest early and decays to neutral ($\alpha_{\text{txt}}\!\to\!1$) late, improving prompt faithfulness without over-constraining details.

For reference-image that requires identity stability, we apply a small late-stage boost to visual cross-attention:
\vspace{-2pt}
\begin{equation}
\tilde S_{v\to v}(u)=\alpha_{\text{img}}(u)\,S_{v\to v},
\end{equation}
where $S_{v\to v}$ represents visual cross-attention scores and
\begin{equation}
\label{factor2}
\alpha_{\text{img}}(u) = 
\begin{cases}
1, & u \in [0,\, 0.6], \\[6pt]
1 + \dfrac{\lambda_{\text{img}}}{2}\!\left(1 - \cos\!\left(\dfrac{\pi (u-0.6)}{0.4}\right)\right), & u \in (0.6,\, 1],
\end{cases}
\qquad \lambda_{\text{img}} = 0.3.
\end{equation}

\subsection{Pyramid Reflection for understanding}

\noindent\textbf{Formulation.}
We cast video question answering as test-time reinforcement learning over a small, ordered evidence set. 
The state at round $r$ is $(s_r, W_r, C_g)$, where $s_r$ is a short search text, $W_r$ is an ordered subset of frames, and $C_g$ is a global caption distilled once from uniformly sampled seeds. 
The action is to reconfigure $W_r$ given $s_r$, either by adding frames (expand) or by pruning to a diverse core (shrink). 
The policy $\pi_s$ is a retrieval rule driven by text–image similarity and a diversity term; it maps $s$ to a distribution over frame indices.
The environment returns an answer $a$ produced by the Actor and a scalar reward $r\in[0,1]$ from the Evaluator. 
Policy improvement is carried out verbally: the Reflector emits a refined $s_{r+1}$ that concentrates on disambiguating cues such as before/after, first/last, motion phase, color, or role. 
The loop stops early when $r$ exceeds a confidence threshold.

\medskip\noindent\textbf{Policy class.}
We instantiate $\pi_s$ with a cached-embedding retriever. 
All $N$ candidate frames are embedded once by a vision encoder; the text side uses $\phi(s)$.
For expand we add the highest-scoring unseen frames by cosine similarity $\langle \mathbf{v}_i,\phi(s)\rangle$, which suits static questions whose evidence is sparse but distinctive. 
For shrink we start broad to preserve chronology, then apply a Maximal Marginal Relevance objective that balances relevance to $\phi(s)$ and pairwise dissimilarity within $W$, which suits dynamic questions where ordering, repetition, or transitions matter. 
In both regimes $W$ is kept in temporal order so the Actor can compare events across $[t_1\!\rightarrow\!t_k]$ rather than hallucinate transitions.

\medskip\noindent\textbf{Value and critic signals.}
The Evaluator provides a calibrated confidence that serves as a value proxy. 
Its scalar reward $r$ both triggers early stopping and conditions the Reflector. 
When $r$ is low, the Reflector returns a short declarative refinement of $s$ that encodes the suspected failure mode: missing entity, wrong time span, ambiguous referent, or occluded phase.
This verbal update reshapes the retrieval distribution without touching model weights, yielding a form of policy gradient in the space of prompts. Our Pyramid Reflection procedure is summarized in Algorithm~\ref{Pyramid_Reflection_algorithm}, 
and the high-level understanding pipeline is shown in Fig.~\ref{fig:understanding_pipeline}. 
The theoretical details of Pyramid Reflection as test-time RL are provided in Appendix~\ref{app:alg-detail}.

The design achieves efficiency by caching frame embeddings once and reducing exploration to lightweight index updates, while the Actor reasons over compact, temporally ordered evidence with fixed global context to maintain scene priors under tight token budgets. The adaptive routing between expansion and MMR-based shrinking aligns retrieval strategies with question structure, enabling effective temporal reasoning at low computational cost.

\begin{table*}[t]
\caption{T2V performance on VBench-Long \citep{huang2024vbench}. }
\label{tab:generation_table}
\centering
\resizebox{\textwidth}{!}{
\begin{tabular}{lccccccccccc}
\toprule
\multirow{2}{*}{\textbf{Method}} &
\multicolumn{3}{c}{\textbf{Overall Scores}} &
\multicolumn{5}{c}{\textbf{Technical Quality}} &
\multicolumn{2}{c}{\textbf{Aesthetic Quality}} \\
\cmidrule(lr){2-4}\cmidrule(lr){5-9}\cmidrule(lr){10-11}
& Total Score$\uparrow$ & Quality$\uparrow$ & Semantic$\uparrow$ &
Subject$\uparrow$ & Background$\uparrow$ & Temporal$\uparrow$ & Motion$\uparrow$ & Dynamic$\uparrow$ &
Aesthetic$\uparrow$ & Imaging$\uparrow$ \\
\midrule
EasyAnimateV5.1~\citep{easyanimate2024}     & 83.42 & 85.03 & 77.01 & 98.00 & 97.41 & 99.19 & 98.02 & 57.15 & \textbf{69.48} & 68.61 \\
MiniMax-Video-01~\citep{minimax2024video}         & 83.41 & 84.85 & 77.65 & 97.51 & 97.05 & 99.10 & 99.22 & 64.91 & 63.03 & 67.17 \\
Kling 1.6~\citep{kling2025}                & 83.40 & 85.20 & 76.99 & 97.40 & 96.84 & 99.64 & 99.13 & 62.22 & 64.81 & 69.70 \\
Wan2.1-T2V-1.3B~\citep{Wan}         & 83.31 & 85.23 & 76.95 & 97.56 & 97.93 & 99.55 & 98.52 & 65.19 & 65.46 & 67.01 \\
HunyuanVideo~\citep{hunyuanvideo2024}            & 83.24 & 85.86 & 75.82 & 97.32 & \textbf{97.93} & 99.49 & 98.99 & \textbf{70.83} & 60.36 & 67.56 \\
Gen-3~\citep{runway2024gen3}                  & 82.32 & 84.11 & 75.17 & 97.01 & 96.62 & 99.61 & 99.23 & 60.14 & 63.34 & 66.82 \\
Vchitect-2.0 (VEnhancer)~\citep{vchitect2025}   & 82.24 & 83.54 & 77.06 & 96.83 & 96.66 & 98.97 & 98.98 & 63.89 & 60.41 & 65.35 \\
CogVideoX1.5-5B~\citep{cogvideox2024}        & 82.17 & 82.78 & 79.76 & 96.87 & 97.35 & 98.88 & 98.31 & 50.93 & 62.79 & 65.02 \\
\midrule
\textbf{UniVid (Ours)}               & \textbf{85.27} & \textbf{86.44} & \textbf{80.58} & \textbf{98.96} & 97.76 & \textbf{99.88} & \textbf{99.25} & 61.83 & 64.21 & \textbf{73.03} \\
\bottomrule
\end{tabular}}

\resizebox{\textwidth}{!}{
\begin{tabular}{lccccccccc}
\toprule
\multirow{2}{*}{\textbf{Method}} & \multicolumn{9}{c}{\textbf{Semantic Fidelity}} \\
\cmidrule(lr){2-10}
& Object$\uparrow$ & Multi-Obj$\uparrow$ & Action$\uparrow$ & Color$\uparrow$ & Spatial$\uparrow$ & Scene$\uparrow$ & Appearance$\uparrow$ & Temporal$\uparrow$ & Overall$\uparrow$ \\
\midrule
EasyAnimateV5.1~\citep{easyanimate2024}        & 89.57 & 66.85 & 95.60 & 77.86 & 76.11 & 54.31 & 23.06 & 24.61 & 26.47 \\
MiniMax-Video-01~\citep{minimax2024video}         & \textbf{97.83} & 76.04 & 92.40 & 90.36 & 75.50 & 50.68 & 20.06 & 25.63 & 27.10 \\
Kling 1.6~\citep{kling2025}                & 93.34 & 73.99 & 96.20 & 81.26 & 79.08 & 55.57 & 20.75 & 24.51 & 26.04 \\
Wan2.1-T2V-1.3B~\citep{Wan}         & 88.81 & 74.83 & 94.00 & 82.00 & 73.04 & 41.96 & 21.81 & 23.13 & 25.50 \\
HunyuanVideo~\citep{hunyuanvideo2024}            & 86.10 & 71.66 & 93.42 & 91.60 & 68.09 & 53.69 & 19.80 & 23.89 & 26.44 \\
Gen-3~\citep{runway2024gen3}                    & 87.81 & 53.64 & 96.40 & 80.90 & 65.03 & 54.57 & 24.31 & 24.71 & 26.69 \\
Vchitect-2.0 (VEnhancer)~\citep{vchitect2025} & 86.61 & 68.84 & \textbf{97.20} & 87.04 & 57.55 & \textbf{56.57} & 23.73 & 25.01 & 27.57 \\
CogVideoX1.5-5B~\citep{cogvideox2024}         & 87.47 & 69.65 & 97.20 & 87.55 & 80.25 & 52.91 & \textbf{24.89} & 25.19 & 27.30 \\
\midrule
\textbf{UniVid (Ours)}           & 94.52 & \textbf{77.45} & 94.20 & \textbf{92.10} & \textbf{80.70} & 46.66 & 23.57 & \textbf{25.91} & \textbf{27.60} \\
\bottomrule
\end{tabular}}
\vspace{-0.45cm}
\end{table*}

\section{Experiments}
\begin{table}[t]
\caption{Comparison on four video QA benchmarks~\citep{msrvtt,TGIF,ActivityNet-QA}.}
\label{tab:understanding_table}
\centering
\resizebox{\textwidth}{!}{
\begin{tabular}{l c cc cc cc cc}
\toprule
\multicolumn{10}{c}{\textbf{Video QA Performance}} \\
\midrule
\multirow{2}{*}{\textbf{Method}} & \multirow{2}{*}{\textbf{LLM size}} &
\multicolumn{2}{c}{\textbf{MSVD-QA}} &
\multicolumn{2}{c}{\textbf{MSRVTT-QA}} &
\multicolumn{2}{c}{\textbf{TGIF-QA}} &
\multicolumn{2}{c}{\textbf{ActivityNet-QA}} \\
\cmidrule(lr){3-4}\cmidrule(lr){5-6}\cmidrule(lr){7-8}\cmidrule(lr){9-10}
 &  & Acc$\uparrow$ & Score$\uparrow$ & Acc$\uparrow$ & Score$\uparrow$ & Acc$\uparrow$ & Score$\uparrow$ & Acc$\uparrow$ & Score$\uparrow$ \\
\midrule
FrozenBiLM~\citep{FrozenBiLM2022}   & 1B & 32.2 & --   & 16.8 & --   & 41.0 & --   & 24.7 & --   \\
VideoChat~\citep{Videochat2023}     & 7B & 56.3 & 2.8  & 45.0 & 2.5  & 34.4 & 2.3  & --   & 2.2 \\
LLaMA-Adapter~\citep{LLaMA-Adapter2024} & 7B & 54.9 & 3.1  & 43.8 & 2.7  & --   & --   & 34.2 & 2.7 \\
Video-LLAMA~\citep{Video-LLAMA2023} & 7B & 51.6 & 2.5  & 29.6 & 1.8  & --   & --   & 12.4 & 1.1 \\
Video-ChatGPT~\citep{VideoChatGPT2024} & 7B & 64.9 & 3.3  & 49.3 & 2.8  & 51.4 & 3.0  & 35.2 & 2.7 \\
Chat-UniVi~\citep{Chat-Univi2024}   & 7B & 65.0 & 3.6  & 54.6 & 3.1  & 60.3 & 3.4  & 45.8 & 3.2 \\
Video-LLaVA~\citep{Video-LLava}     & 7B & 70.7 & 3.9 & 59.2 & 3.5 & 70.0 & 4.0 & 45.3 & 3.3 \\
BT-Adapter~\citep{BT-Adapter2024}   & 7B & 67.5 & 3.7  & 57.0 & 3.2  & --   & --   & 45.7 & 3.2 \\
Valley-v3~\citep{Valley-v3}         & 7B & 60.5 & 3.3  & 51.1 & 2.9  & --   & --   & 45.1 & 3.2 \\
FreeVA~\citep{FreeVA}           & 7B & 73.8 & 4.1  & 60.0 & 3.5  & --   & --   & 51.2 & 3.5 \\
DeepStack-L~\citep{DeepStack-L}  & 7B & 76.0 & 4.0  & --   & --   & --   & --   & 49.3 & 3.1 \\
IG-VLM (LLaVA-v1.6)~\citep{IG-VLM} & 7B & 78.8 & 4.1  & 63.7 & 3.5  & --   & 4.0  & 54.3 & 3.4 \\
SF-LLaVA-7B~\citep{SF-LLaVA-7B} & 7B & 79.1 & 4.1 & \textbf{65.8} & \textbf{3.6} & \textbf{78.7} & \textbf{4.2} & 55.5 & 3.4 \\
\bottomrule
\textbf{UniVid (Ours)} & 7B & \textbf{80.1} & \textbf{4.2} & 61.4 & 3.4 & 75.0 & 4.1 & \textbf{58.8} & \textbf{3.6} \\
\midrule
\end{tabular}
}
\vspace{-0.6cm}
\end{table}

\subsection{Dataset and Metrics}

\paragraph{Datasets.}
We evaluate UniVid on established benchmarks for both video generation and understanding. For generation, we train on curated samples from OpenVid-1M, a large-scale text-to-video dataset, and evaluate on VBench, a comprehensive benchmark suite for video generative models that provides fine-grained evaluation metrics across multiple dimensions. For understanding, we train on 20k samples from the ActivityNet-QA train dataset ~\citep{ActivityNet-QA} and evaluate on four comprehensive video QA benchmarks: MSVD-QA ~\citep{msrvtt} with 1,970 video clips and 50.5K QA pairs, MSRVTT-QA ~\citep{msrvtt} with 10K videos, 243K QA pairs, TGIF-QA ~\citep{TGIF} containing 165K QA pairs for animated GIFs, and the ActivityNet-QA test dataset ~\citep{ActivityNet-QA} with 58,000 QA pairs on 5,800 complex web videos. These datasets cover diverse temporal reasoning scenarios across short to medium-length video clips, ranging from brief animated sequences to multi-minute activity videos. 

\paragraph{Evaluation metrics.}
For video generation, we evaluate on VBench across multiple fine-grained dimensions: Technical Quality metrics including Subject consistency, Background preservation, Temporal flickering, Motion smoothness, and Dynamic degree; Aesthetic Quality measures covering overall visual appeal and imaging quality; and Semantic Fidelity metrics assessing Object accuracy, Multi-object handling, Action fidelity, Color accuracy, Spatial relationships, Scene consistency, Appearance preservation, and Temporal coherence. For video understanding, we report average accuracy and scores on each benchmark dataset. 

\subsection{Implementation Details}

We adopt a three-stage hierarchical training recipe. It initializes UniVid from strong public checkpoints to reduce compute. For generation, we couple the BAGEL-7B~\citep{BAGEL} with Wan 2.2 5B TI2V model~\citep{Wan} via a textual adapter and LoRA on DiT~\citep{DBLP:conf/iccv/PeeblesX23}, keeping other weights frozen. For understanding, we tune only the connector and the last two ViT blocks on ActivityNet QA~\citep{ActivityNet-QA} with dialog style supervision while the LLM remains frozen. Finally, we co-train both tasks to refine the connector and obtain additive gains. Sequence parallelism enables long high-resolution clips. For details, see Appendix \ref{app:post}.

For generation, we use a flow-matching ODE sampler with classifier-free guidance and a universal negative prompt. Unless noted, videos are sampled at \(1280 \times 704\) resolution, 121 frames at 24 fps; the guidance scale is set to 5.0 for both T2V and I2V with 50 inference steps. At input time, the LLM receives the text prompt together with image ViT embeddings and VAE latents; it outputs conditional textual tokens. During generation, Wan 2.2 consumes these conditional textual tokens and image via cross-attention. Our Temperature Modality Alignment schedule applies a cosine-scheduled text gain that transitions from \(\alpha_{\text{txt}} = 1.3\) to 1.0 over the first 40\% of denoising steps (\(u \in [0, 0.4]\)), then maintains \(\alpha_{\text{txt}} = 1.0\) for the remaining steps. This enhances text guidance during early denoising when structural decisions are made, while allowing finer details to emerge in later stages.

For understanding, we uniformly sample a pool of \(N=64\) frames per video and cache their SigLIP2 image embeddings; subsequent selection reuses cached features.
Global context is a caption summarized from 16 uniformly spaced seed frames.
Query–image ranking uses SigLIP2 cosine similarity with L2-normalized features and batch size \(64\).
Static questions follow a \(4 \rightarrow 8 \rightarrow 16\) keyframe schedule.
Dynamic questions follow \(64 \rightarrow 32 \rightarrow 16\) with MMR down-selection, \(\lambda=0.5\). Confidence is accepted when the Evaluator’s score is at least \(0.7\) or the verdict is accept, with at most \(R \le 3\) rounds. The LLM determines routing between static and dynamic modes. For implementation, we use DeepSeek v3.1 to serve as the Evaluator and determine the type of questions and Qwen-plus to serve as the Reflector. Full prompt texts are listed in the Appendix ~\ref{sec:prompts}.

\subsection{Main Results}

\paragraph{Generation quantitative results.}
We evaluate UniVid on the challenging VBench-Long benchmark~\citep{huang2024vbench}. As shown in Tab.~\ref{tab:generation_table}, UniVid establishes a new state of the art with an overall score of 85.27, outperforming prior leading systems such as EasyAnimateV5.1~\citep{easyanimate2024}, MiniMax-Video-01~\citep{minimax2024video}, and Kling 1.6~\citep{kling2025}. In particular, UniVid exhibits clear advantages in semantic alignment (80.58), highlighting its superior capability in faithfully rendering objects, actions, and multi-object interactions. On the technical side, it attains near-perfect temporal (99.88) and motion (99.25) consistency, validating the effectiveness of our long-context dynamics module. Moreover, UniVid delivers the best imaging score (73.03), reflecting sharper details and more stable visual quality compared with prior systems, as shown in Fig.~\ref{fig:teaser}, which demonstrates high-quality visual generation.

Beyond overall scores, UniVid demonstrates consistent gains in semantic fidelity. As summarized in the Semantic Fidelity block of Tab.~\ref{tab:generation_table}, it achieves leading results on multi-object reasoning (77.45), color faithfulness (92.10), and spatial grounding (80.70), while remaining competitive in action depiction and appearance consistency. These improvements suggest that our design choices—particularly the integration of hierarchical scene representation with dynamic frame alignment—substantially enhance both controllability and alignment with textual prompts. Taken together, the results indicate that UniVid pushes forward the frontier of long-horizon text-to-video generation by simultaneously ensuring high-fidelity semantics and strong technical as well as aesthetic quality. More examples of video generation can be seen in Appendix ~\ref{sec:example}.

\begin{figure}[t] %
  \centering
  \includegraphics[width=\columnwidth]{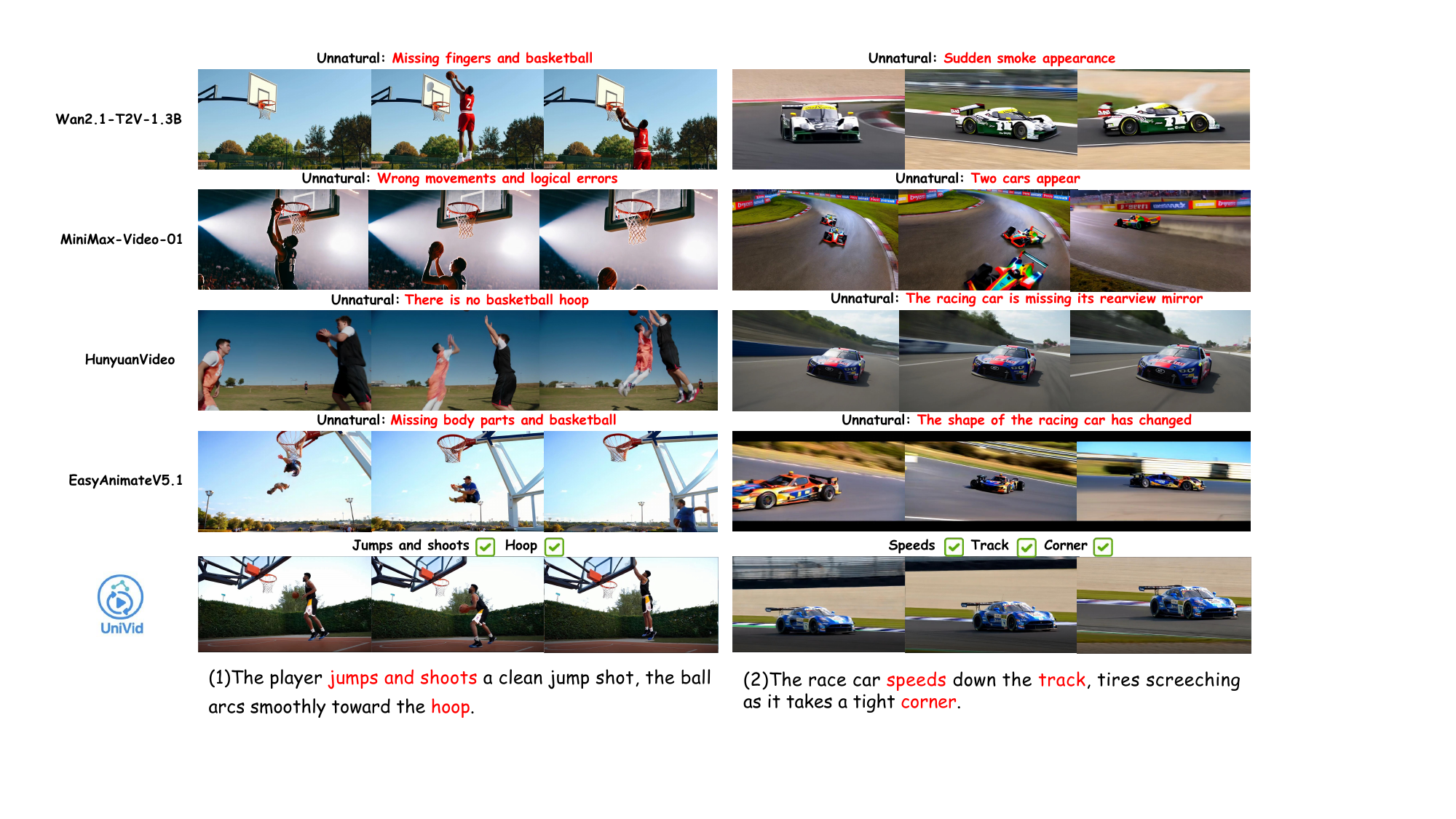}
  \caption{Comparisons with State-of-the-Art Video Generation Models~\citep{Wan,minimax2024video,hunyuanvideo2024,easyanimate2024}.}
  \label{fig:generation_comparasion}
  \vspace{-0.5cm}
\end{figure}

\paragraph{Generation qualitative results.}
Fig.~\ref{fig:generation_comparasion} compares UniVid with Wan2.1-T2V-1.3B~\citep{Wan}, MiniMax-Video-01~\citep{minimax2024video}, HunyuanVideo~\citep{hunyuanvideo2024}, and EasyAnimateV5.1~\citep{easyanimate2024}. Competing models often show missing basketballs or distorted cars, while UniVid generates coherent jump shots and realistic racing scenes with stable dynamics and faithful semantics.

\paragraph{Understanding quantitative evaluation.}
Across MSVD-QA \citep{msrvtt}, MSRVTT-QA \citep{msrvtt}, TGIF-QA \citep{TGIF}, and ActivityNet-QA \citep{ActivityNet-QA}, UniVid sets the 7B-scale state of the art on MSVD-QA and ActivityNet-QA and remains competitive on the other two (Tab.~\ref{tab:understanding_table}), despite a smaller post-training set and no test-time ensembling. Joint finetuning of generation and understanding with Pyramid Reflection strengthens the abilities these datasets emphasize: better action–entity binding and object or attribute grounding in short open-domain clips, stronger temporal reasoning over frame sequences, and more reliable long-range evidence retrieval in untrimmed videos.

As illustrated before, UniVid performs robust multi-frame reasoning with our Pyramid Reflection loop. Starting from a global caption and automatic type detection, the system first produces an initial answer, which is then scored by the evaluator; when evidence is insufficient, the reflector issues a refined, declarative query that re-ranks keyframes toward the true scene. This Pyramid Reflection steers attention from opening credits to the lane shots, yielding a consistent interpretation of roles (in the example of Fig.~\ref{fig:understanding_pipeline}: bowler and nearby teammate/coach) grounded in the visual context rather than spurious cues. The dynamic keyframe schedule reduces the number of inspected frames while maintaining accuracy, demonstrating both evidence tracing and efficiency gains in short-clip understanding. More examples of video understanding can be seen in Appendix ~\ref{sec:example}.

\begin{figure}[t]
  \centering
  \includegraphics[width=\linewidth]{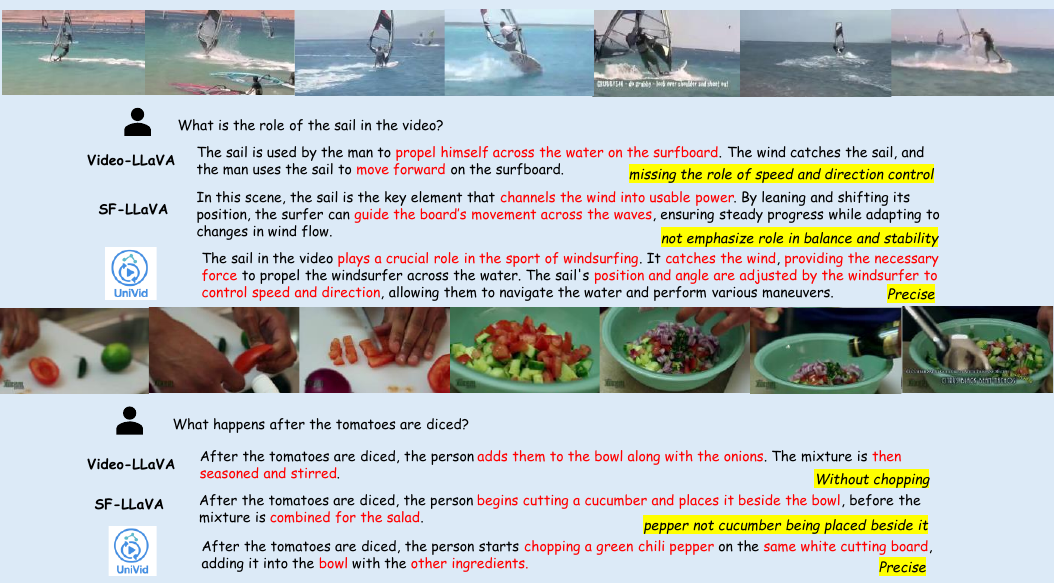}
  \caption{Comparisons of State-of-the-Art Video Understanding Models ~\citep{Video-LLava,SF-LLaVA-7B}.}
  \label{fig:understand_comparasion}
  \vspace{-0.5cm}
\end{figure}

\paragraph{Understanding qualitative results.}
We compare UniVid with Video-LLaVA~\citep{Video-LLava} and SF-LLaVA ~\citep{SF-LLaVA-7B} on video QA; as shown in Fig.~\ref{fig:understand_comparasion}, baselines often give plausible but incomplete statements. These examples highlight UniVid’s stronger action–entity binding, temporal reasoning, and resistance to distractor frames, yielding precise and concise answers.
Additionally, we conduct systematic ablation experiments to validate the contributions of UniVid. The results and analyses are provided in the Appendix ~\ref{ablation}.

\section{Conclusion}
We introduced UniVid, a unified video model that couples an MLLM with a diffusion decoder via a lightweight conditioning adapter to both understand and generate videos. Two key mechanisms enable this: Temperature Modality Alignment schedules cross-modal attention across flow steps to preserve prompt faithfulness while refining details, and Pyramid Reflection performs query-driven keyframe selection for efficient temporal reasoning. With these components, UniVid achieves state-of-the-art or competitive results on VBench-Long and multiple video-QA benchmarks while avoiding costly retraining of image-centric backbones. We release UniVid to support research on practical, controllable, and truly unified video intelligence.

\clearpage

\input{iclr2026_conference.bbl}
\clearpage
\input{appendix}

\end{document}

%% file: math_commands.tex
\usepackage{amsmath,amsfonts,bm}

\def\eqref#1{equation~\ref{#1}}

\def\1{\bm{1}}

\DeclareMathAlphabet{\mathsfit}{\encodingdefault}{\sfdefault}{m}{sl}
\SetMathAlphabet{\mathsfit}{bold}{\encodingdefault}{\sfdefault}{bx}{n}

%% file: appendix.tex
\appendix
\section{Appendix}

\subsection{LLM Use Declaration}
Large Language Models (ChatGPT) were used exclusively to improve the clarity and fluency of English writing. They were not involved in research ideation, experimental design, data analysis, or interpretation. The authors take full responsibility for all content.

\subsection{Hierarchical Post Training}
\label{app:post}
\paragraph{Initialization.}
To avoid the prohibitive cost of training a unified video model from scratch, we bootstrap UniVid from strong, publicly available checkpoints and finetune only small subsets of parameters. Our architecture follows the BAGEL~\citep{BAGEL} design framework, adopting its multimodal integration approach with three key components: Qwen2~\citep{Qwen2} as the LLM backbone with standard architectural choices such as RMSNorm~\citep{RMSNorm}, SwiGLU~\citep{GLU}, RoPE~\citep{RoPE}, GQA~\citep{GQA}, and QK-Norm~\citep{QKNorm} for training stability, SigLIP2-so400m/14~\citep{SigLip2} as the ViT~\citep{vit} encoder for visual understanding with NaViT support for native aspect ratios, and a pre-trained FLUX VAE with 8× downsampling and frozen weights. The framework interleaves text, ViT, and VAE tokens within the LLM using generalized causal attention, where tokens attend to all preceding modality splits while maintaining appropriate attention patterns within each modality.

\paragraph{Data curation and formatting.}
For understanding, we align our data format with the dialog style used by Video-ChatGPT~\citep{VideoChatGPT2024}. ActivityNet-QA annotations \((\texttt{video\_id}, q, a)\) are converted into structured conversations. Specifically, each sample is represented as a JSON object containing three fields:
(1) an identifier, (2) a video reference, and (3) a conversations array consisting of two turns, a user query and the corresponding model response. For generation, we curate a subset of OpenVid-1M to form text/image to video pairs. Videos are uniformly sub-sampled and preprocessed identically to inference.

\paragraph{Stage I generation branch alignment.}
We couple the MLLM with Wan 2.2 and adapt the conditioning path so that MLLM-produced tokens can reliably steer synthesis. Concretely, we (i) insert a textual adapter between the LLM tokens, with dynamic sequence length adaptation, and (ii) apply LoRA to the DiT cross-attention layers; all other DiT/MLLM weights remain frozen. Training uses a standard flow-matching objective with classifier-free guidance dropout on text, optimizing only the context projector and LoRA parameters. This stage preserves MLLM's native understanding while aligning Wan's generation to the rich semantics emitted by MLLM.

\paragraph{Stage II understanding adaptation.}
We finetune for video QA on ActivityNet-QA using 20k samples from the dataset. Each sample concatenates the question with a \texttt{<video>} placeholder, and we feed a multi-frame clip obtained by uniform sampling. Frames are encoded by the ViT into visual tokens and projected to the LLM space via the connector. We adopt instruction SFT for video: compute autoregressive cross-entropy only on the assistant turns; user tokens are fully masked to prevent label leakage. To keep compute moderate while injecting temporal cues, we finetune only the last two ViT blocks and the connector for 4 epochs, keeping the LLM frozen. 

\paragraph{Stage III joint training.}
Finally we co-train generation and understanding to let the two branches benefit from each other. We alternate mini-batches from the two tasks and update primarily the connector within the MLLM and the textual adapter which bridges MLLM outputs to Wan 2.2, using the sum of the velocity loss for generation and the masked autoregressive loss for understanding. This joint stage optimizes the MLLM to improve both understanding and generation performance.

\subsection{More examples of Video generation and understanding. }
\label{sec:example}
\begin{figure}[t] 
  \centering
  \includegraphics[width=\columnwidth]{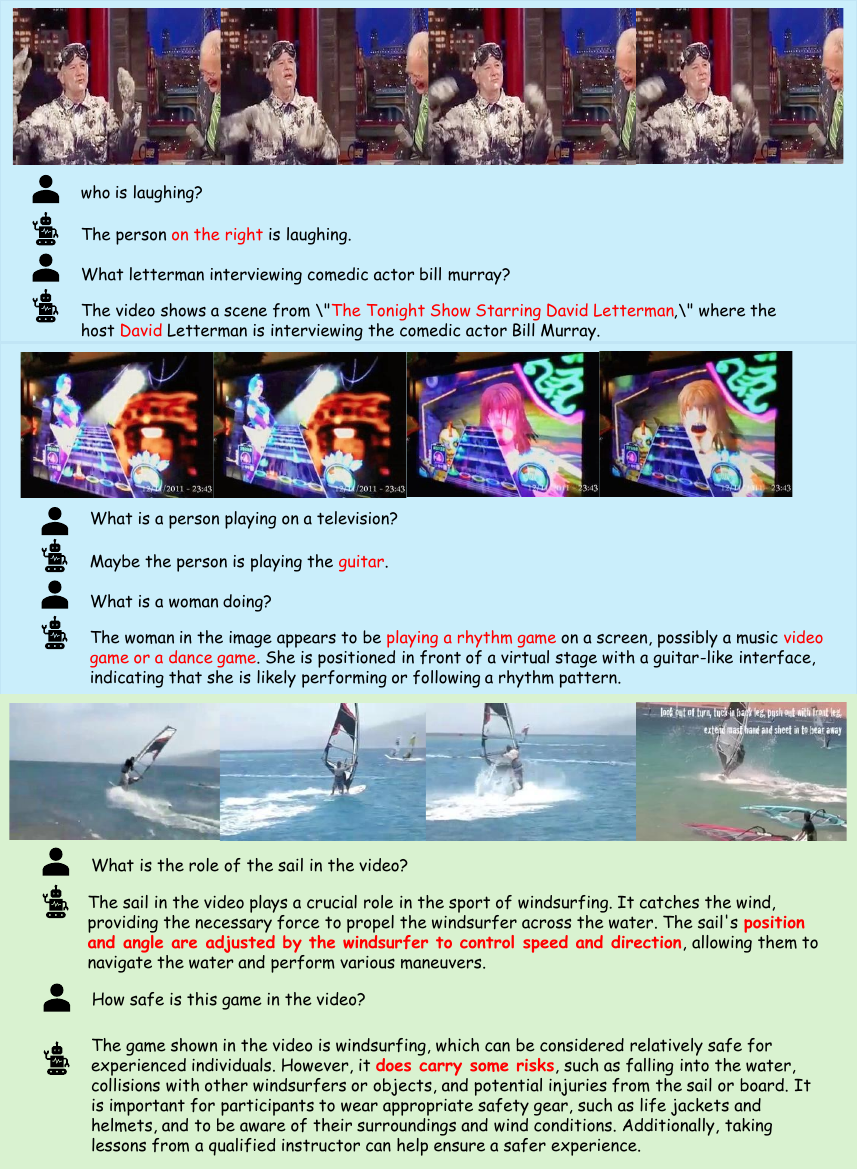}
  \caption{The qualitative results of the video understanding. Blue for static questions, green for dynamic questions.}
  \label{fig:und_ qualitative}
\end{figure}

\begin{figure}[t]
  \centering
  \includegraphics[width=\columnwidth]{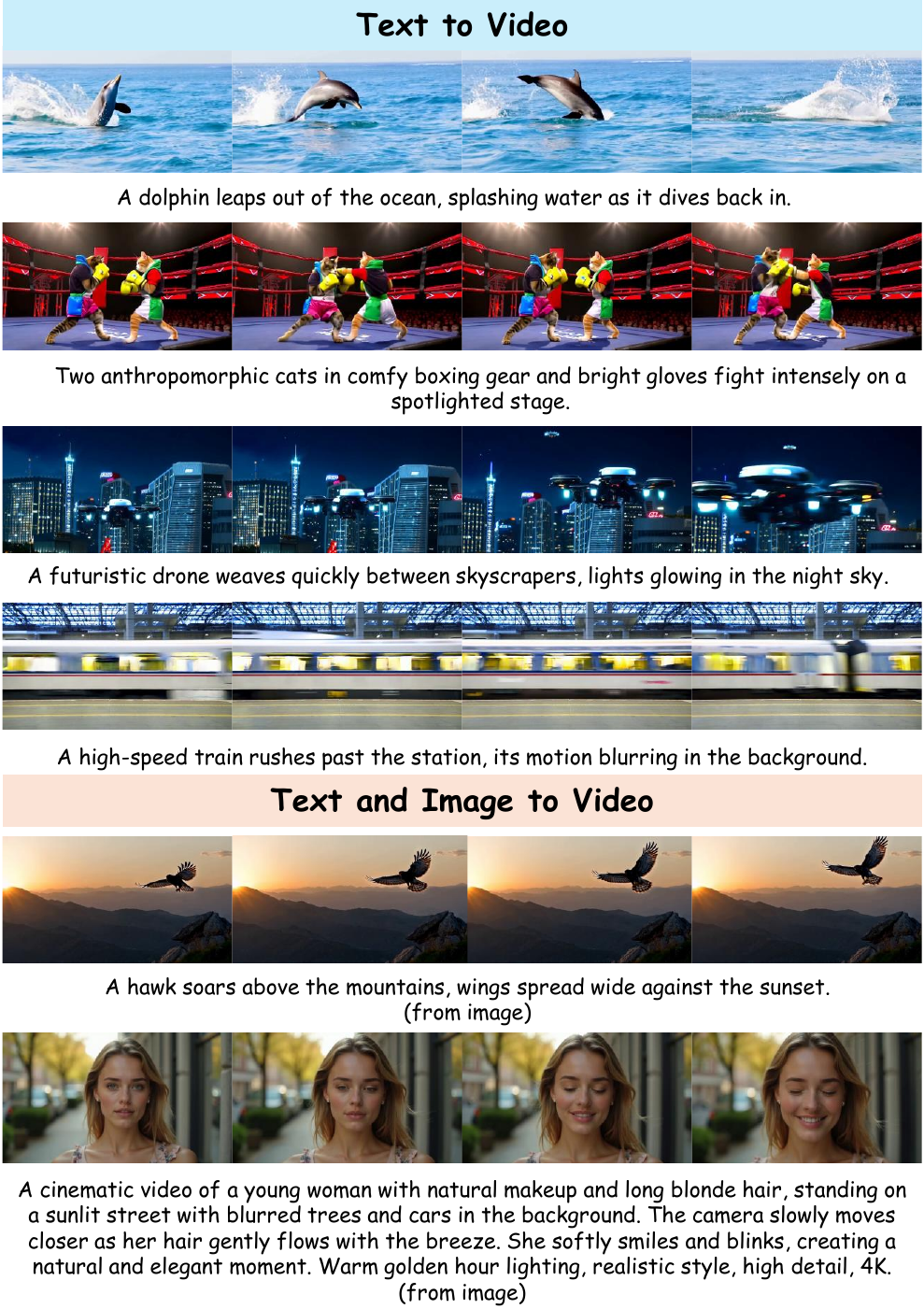}
  \caption{The qualitative results of T2V and TI2V generation.}
  \label{fig:gen_qualitative}
\end{figure}

We provide more examples of video understanding and generation in Fig.~\ref{fig:und_ qualitative} and Fig. ~\ref{fig:gen_qualitative}

\clearpage
\subsection{Text Prompts used in the understanding}

\label{sec:prompts}

\begin{figure}[t] %
  \centering
  \includegraphics[width=\columnwidth]{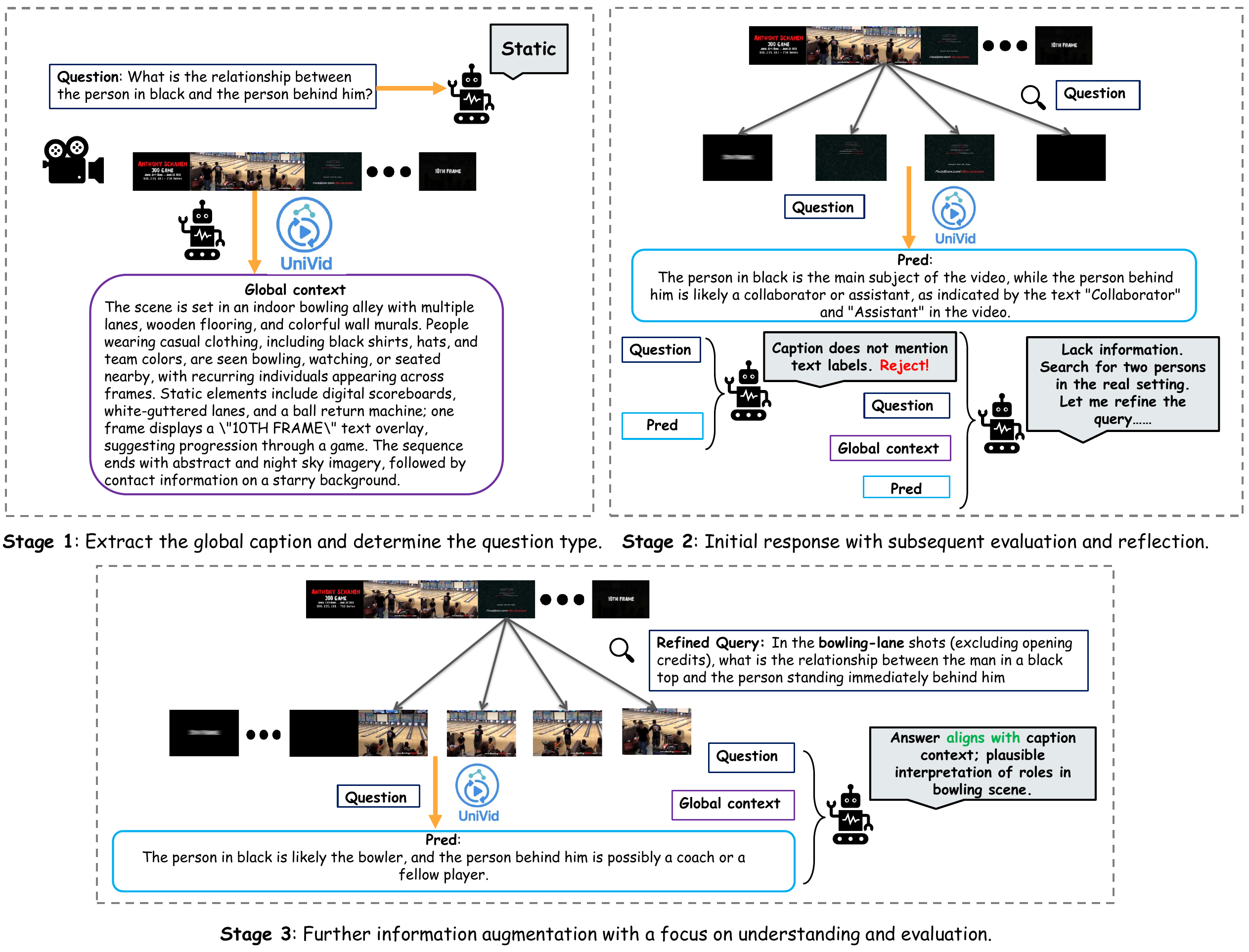}
  \caption{The pipeline of the video understanding.}
  \label{fig:understanding_pipeline}
\end{figure}

\begin{tcolorbox}[promptbox, before skip=0pt, after skip=0pt]
\textbf{Role.} Classify a video question as \texttt{static} or \texttt{dynamic}. Output JSON only.\\
\textbf{Definitions.} 
\begin{itemize}[itemsep=1pt,leftmargin=10pt]
\item \texttt{dynamic}: requires temporal reasoning such as counting, repetition, order, or changes over time (e.g., ``how many times'', ``before/after'', ``first/last'').
\item \texttt{static}: can be answered from a small set of unordered frames (identity, attribute, location, scene, one-shot action).
\end{itemize}
\textbf{Question.} \emph{\{question\}}\\
\textbf{Return.} Single-line JSON with fields: \texttt{qtype} (\texttt{"static"} or \texttt{"dynamic"}), \texttt{rationale} (1–2 short phrases; no extra text).
\end{tcolorbox}

\begin{center}
\textbf{1: Question Type Classification Prompt}
\end{center}

\begin{tcolorbox}[promptbox]
\textbf{Role.} Summarize chronologically ordered frame notes into a compact global caption. Do not invent facts.\\
\textbf{Input.} Frame-wise notes (earlier $\rightarrow$ later):\\
\emph{- \{note\_1\}\\ - \{note\_2\}\\ \dots}\\
\textbf{Write.} One global caption (2–4 sentences) that connects multiple frames, focusing on: 
(1) moving entities with consistent appearance and actions across time; 
(2) static scene objects and their states; 
(3) temporal hints only if explicitly evidenced (e.g., “then”, “later”, “repeatedly”). 
Style: terse and factual; no bullet lists, storytelling, or frame-by-frame recitation.
\end{tcolorbox}

\begin{center}
\textbf{2: Frame Summarization Prompt}
\end{center}

\begin{tcolorbox}[promptbox]
\textbf{Role.} Precise evaluator for video-QA. Return a \emph{single-line} JSON only (no Markdown/code).\\
\textbf{Keys.} 
\texttt{score} (float 0..1), 
\texttt{verdict} (\texttt{"accept"} if score $\ge$ 0.7 else \texttt{"reject"}), 
\texttt{brief\_reason} (1–2 short bullets).\\
\textbf{Example user.} \emph{\{one\_shot\_user\}}\\
\textbf{Example assistant.} \emph{\{one\_shot\_assistant\}}\\
\textbf{Your task.} Given the current case, output the JSON only.
\end{tcolorbox}

\begin{center}
\textbf{3: Answer Evaluation Prompt}
\end{center}

\begin{tcolorbox}[promptbox]
\textbf{Role.} Reflector in a video-understanding pipeline. You receive the question, a global caption (from 16 uniformly sampled frames), the last answer (low confidence/rejected), and its evaluation JSON.\\
\textbf{Objective.} Analyze why the answer likely fails (missing object, wrong span, ambiguity, etc.) and produce a single short \emph{declarative} retrieval text for the next round of keyframe selection.\\
\textbf{Strict rules.}
(1) Output JSON only with key \texttt{refined\_query}. 
(2) \texttt{refined\_query} $\le$ 25 tokens, declarative statement (not a question), capturing disambiguating cues (entities, attributes, actions, temporal hints, viewpoint). 
(3) If confidence is already good (score $\ge 0.7$ or verdict=\texttt{"accept"}), return an empty string. 
(4) Prefer concrete visual cues (colors, clothing, object names, motion phase, timestamps, left/right, first/last). 
(5) No speculation or unseen entities.\\
\textbf{Inputs.} 
Question: \emph{\{question\}} \quad
Global caption: \emph{\{global\_caption\}} \quad
Last answer: \emph{\{last\_answer\}} \quad
Evaluation JSON: \emph{\{eval\_json\}}\\
\textbf{Return.} \texttt{\{"refined\_query": "..."\}}
\end{tcolorbox}

\begin{center}
\textbf{4: Reflection Prompt}
\end{center}

\begin{tcolorbox}[promptbox]
\textbf{Role.} Assist video understanding via per-frame analysis. Describe the main objects and actions in \emph{this single frame} concisely.\\
\textbf{Focus.}
(1) Living entities: distinct entities (appearance, clothing, color, species), likely roles, and what each is doing (verb phrases). 
(2) Static objects \& scene: salient items and states (color, shape, on/off, open/closed, broken/intact), plus scene context (indoor/outdoor, location hints).\\
\textbf{Style.} Specific but brief; no speculation; 2–4 short sentences.
\end{tcolorbox}

\begin{center}
\textbf{5: Single-Frame Analysis Prompt}
\end{center}

\begin{tcolorbox}[promptbox]
\textbf{Role.} Answer concisely using only the question and the global video caption.\\
\textbf{Inputs.} 
Question: \emph{\{question\}} \quad
Global caption (may miss fine details): \emph{\{global\_caption\}}\\
\textbf{Instruction.} Produce one short answer (1–2 sentences). If information is insufficient, reply: \emph{“Not enough evidence from global caption.”}
\end{tcolorbox}

\begin{center}
\textbf{6: Global Answer Prompt}
\end{center}

\subsection{Pyramid Reflection as Test-time RL}
\label{app:alg-detail}

We cast Pyramid Reflection as a test-time reinforcement learning procedure operating on an ordered evidence set. At round $r$, the state is $x_r=(s_r,W_r,C_g)$, where $s_r$ is a short search text, $W_r$ is the ordered working set of frames, and $C_g$ is a global caption distilled once from uniformly sampled seeds. The action reconfigures $W_r$ given $s_r$ via an expand or shrink policy. The Actor answers from $(W_r,C_g)$, and the Evaluator returns a score $R_r\in[0,1]$ and a verdict that controls early stopping. All frame embeddings are computed once and cached; later rounds update indices and similarity or diversity scores only.

Frame selection uses a vision–language retriever with cosine similarity. Let $\phi(s)$ be the text embedding and $\{\mathbf v_i\}_{i=1}^N$ the cached frame embeddings:
\begin{equation}
\mathrm{sim}(i,s)=\big\langle \widehat{\mathbf v}_i,\widehat{\phi(s)}\big\rangle .
\label{eq:sim}
\end{equation}
We define a soft retrieval policy over the pool $P$:
\begin{equation}
\pi(i\mid s)=
\frac{\exp\big(\mathrm{sim}(i,s)/\tau\big)}
     {\sum_{j\in P}\exp\big(\mathrm{sim}(j,s)/\tau\big)} .
\label{eq:policy}
\end{equation}
Sampling sequentially without replacement with joint probability $\prod_{\ell=1}^{K}\pi(i_\ell\mid s, i_{<\ell})$ and respecting chronology yields $W_s$.

In the expand mode, at target size $K_t$ we add the top $m$ unseen frames by similarity (no duplicates):
\begin{equation}
\Delta_t=\underset{i\in P\setminus S_{\mathrm{sel}}}{\arg\max}^{\,m}\ \mathrm{sim}(i,s_{t-1}),
\qquad
S_{\mathrm{sel}}\leftarrow S_{\mathrm{sel}}\cup \Delta_t,\quad
m=K_t-|S_{\mathrm{sel}}|.
\label{eq:expand}
\end{equation}
In the shrink mode, with current $S_{\mathrm{sel}}$ and target $K_t\in\{32,16\}$, we apply Maximal Marginal Relevance:
\begin{equation}
S_{\mathrm{sel}}
=\underset{S\subseteq S_{\mathrm{sel}},\,|S|=K_t}{\arg\max}
\sum_{i\in S}\Big[\lambda\,\mathrm{sim}(i,s_{t-1})
-(1-\lambda)\max_{j\in S\setminus\{i\}}\mathrm{sim}(i,j)\Big] .
\label{eq:mmr}
\end{equation}

\begin{table*}[t]
\caption{Ablation study of UniVid on VBench-Long. \textit{w/o} means ``without''. Best results are \textbf{bold}.}
\label{tab:video_ablation}
\centering
\resizebox{\textwidth}{!}{
\begin{tabular}{lccccccccccc}
\toprule
\multirow{2}{*}{\textbf{Model}} &
\multicolumn{3}{c}{\textbf{Overall Scores}} &
\multicolumn{5}{c}{\textbf{Technical Quality}} &
\multicolumn{2}{c}{\textbf{Aesthetic Quality}} \\
\cmidrule(lr){2-4}\cmidrule(lr){5-9}\cmidrule(lr){10-11}
& Total Score$\uparrow$ & Quality$\uparrow$ & Semantic$\uparrow$ &
Subject$\uparrow$ & Background$\uparrow$ & Temporal$\uparrow$ & Motion$\uparrow$ & Dynamic$\uparrow$ &
Aesthetic$\uparrow$ & Imaging$\uparrow$ \\
\midrule
UniVid (base)     & 76.25 & 77.11 & 72.82 & 93.82 & 93.43 & 94.15 & 94.04 & 57.16 & 58.47 & 65.65 \\
UniVid (w/o MLLM)         & 77.82 & 78.69 & 74.32 & 94.55 & 94.78 & 95.19 & 94.79 & 58.08 & 59.88 & 66.01 \\
UniVid (w/o TMA)         & 80.42 & 81.51 & 76.04 & 96.55 & 95.91 & 97.12 & 96.25 & 59.98 & 62.08 & 67.10 \\

\midrule
\textbf{UniVid (Full)}               & \textbf{85.27} & \textbf{86.44} & \textbf{80.58} & \textbf{98.96} & \textbf{97.76} & \textbf{99.88} & \textbf{99.25} & \textbf{61.83} & \textbf{64.21} & \textbf{73.03} \\
\bottomrule
\end{tabular}}

\resizebox{\textwidth}{!}{
\begin{tabular}{lccccccccc}
\toprule
\multirow{2}{*}{\textbf{Model}} & \multicolumn{9}{c}{\textbf{Semantic Fidelity}} \\
\cmidrule(lr){2-10}
& Object$\uparrow$ & Multi-Obj$\uparrow$ & Action$\uparrow$ & Color$\uparrow$ & Spatial$\uparrow$ & Scene$\uparrow$ & Appearance$\uparrow$ & Temporal$\uparrow$ & Overall$\uparrow$ \\
\midrule
UniVid (base)        & 89.53 & 73.32 & 89.41  & 87.86 & 76.13 & 42.32 & 19.03 & 21.60 & 22.48 \\
UniVid (w/o MLLM)       & 90.80 & 74.37 & 90.12 & 87.99 & 76.63 & 43.32 & 20.57 & 22.26 & 22.98 \\
UniVid (w/o TMA)       & 91.51 & 75.42 & 91.53 & 89.33 & 77.58 & 44.61 & 21.03 & 23.62 & 24.13 \\
\midrule
\textbf{UniVid (Full)}           & \textbf{94.52} & \textbf{77.45} & \textbf{94.20} & \textbf{92.10} & \textbf{80.70} & \textbf{46.66} & \textbf{23.57} & \textbf{25.91} & \textbf{27.60} \\
\bottomrule
\end{tabular}}
\end{table*}

We adopt a verbal policy–improvement view~\citep{reflexion}. Let the objective be the expected Evaluator value under the retrieval policy:
\begin{equation}
J(s)=\mathbb{E}_{i_{1:K}\sim\pi(\cdot\mid s)}\!\left[V(W_s)\right],
\label{eq:J}
\end{equation}
with
\begin{equation}
V(W_s)=\mathbb{E}\!\left[R\,\big|\,W_s,C_g\right].
\label{eq:V}
\end{equation}
Using the likelihood–ratio identity with a baseline $b$ yields
\begin{equation}
\nabla_s J(s)=
\mathbb{E}\!\left[
\Big(\sum_{t=1}^K \nabla_s \log \pi(i_t\mid s, i_{<t})\Big)\,(R-b)
\right].
\label{eq:reinforce}
\end{equation}
A single ascent step motivates a verbal update to the search text:
\begin{equation}
s_{r+1}= s_r+\eta
\Big(\sum_{t=1}^K \nabla_s \log \pi(i_t\mid s_r, i_{<t})\Big)\,(R_r-b),
\label{eq:update}
\end{equation}
where we use the softmax score function with
$g_i(s):=\nabla_s\mathrm{sim}(i,s)$ and
$\bar g(s):=\mathbb{E}_{j\sim\pi(\cdot\mid s)}g_j(s)$:
$\ \nabla_s\log\pi(i\mid s)=\tau^{-1}\!\big(g_i(s)-\bar g(s)\big)$,
so the edit in $s$ aligns with frames that explain higher return through the text encoder $\phi(\cdot)$. Practically, the reflector inserts temporally and semantically discriminative cues (entities, colors, viewpoints, before/after, first/last, motion phase), which increases $\mathrm{sim}(i,s)$ for diagnostic frames and decreases it for distractors, implementing Eq.~\ref{eq:update} in language space without parameter updates.

To connect the update with both expand and shrink, we use a piecewise-smooth set surrogate that trades relevance against redundancy (subgradients at ties):
\begin{equation}
\tilde V(W_s)=\frac{1}{K}\sum_{i\in W_s}\mathrm{sim}(i,s)\;-\;\gamma\max_{i\neq j\in W_s}\mathrm{sim}(i,j).
\label{eq:surrogate}
\end{equation}
Since $\partial\,\mathrm{sim}(i,s)/\partial s$ points toward $\mathbf v_i$ via $\phi(s)$, the gradient $\nabla_s\tilde V(W_s)$ is aligned with the direction in Eq. \ref{eq:reinforce}. If the reflector’s edit correlates with the advantage $A_r=R_r-b$, then for a sufficiently small step size $\eta$ the expected first-order improvement satisfies
\begin{equation}
\mathbb{E}\!\left[J(s_{r+1})-J(s_r)\right]
\approx
\eta\ \mathbb{E}\!\left[\left\langle
\sum_{t}\nabla_s\log\pi(i_t\mid s_r, i_{<t}),\ s_{r+1}-s_r
\right\rangle A_r\right]\ \ge\ 0 .
\label{eq:improve}
\end{equation}
Early stopping is triggered when the Evaluator score exceeds a fixed threshold:
\begin{equation}
\text{stop at round } r\ \ \text{if}\ \ R_r \ge \tau,\qquad \tau=0.7.
\label{eq:stop}
\end{equation}
With cached features, each round requires only similarity and diversity scoring together with reasoning over a compact, temporally ordered $W_r$, which concentrates the Actor on temporal relations under a tight token budget and improves video understanding with low computational cost.

\begin{figure}[t]
  \centering
  \includegraphics[width=\linewidth]{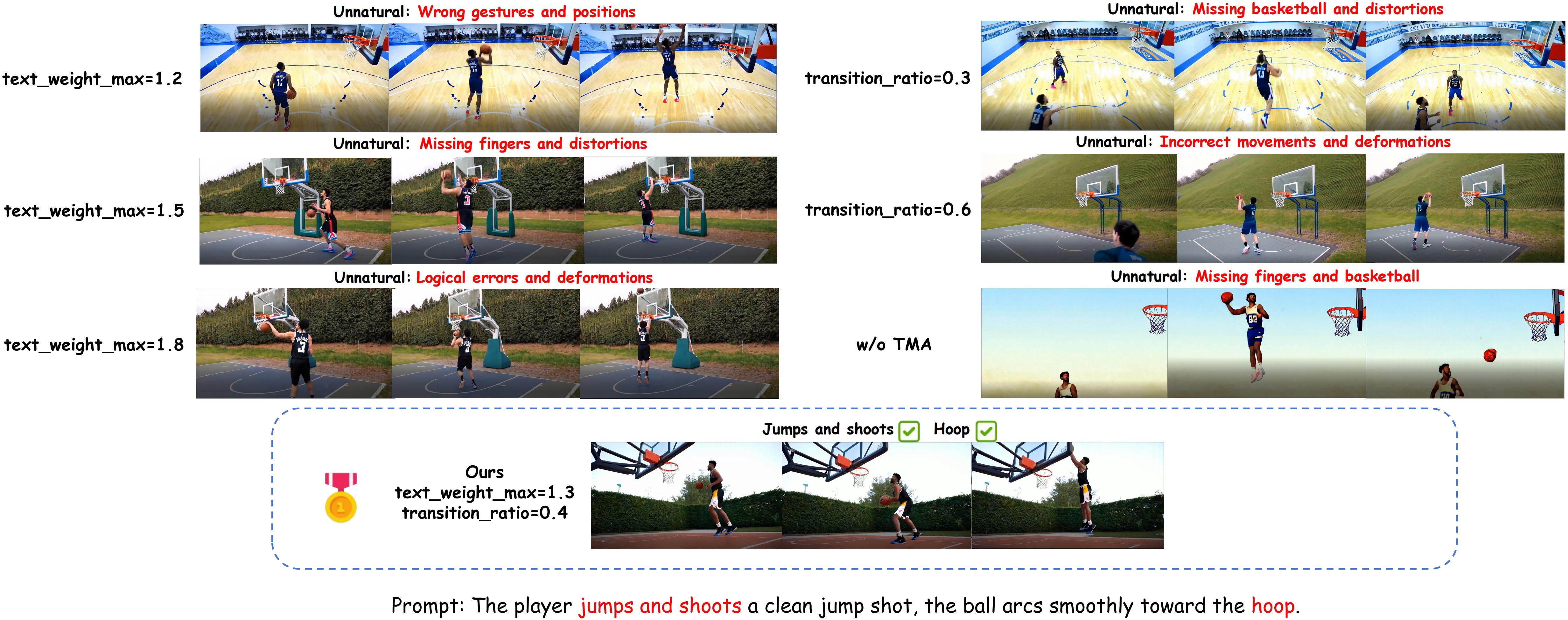}
  \caption{Ablation Study on Temperature Modality Alignment.}
  \label{fig:ablation_tma}
\end{figure}

\begin{table}[t]
\caption{Ablation study of UniVid on four video QA benchmarks. Acc. denotes accuracy (\%), Score denotes average rating (0--5). \textit{w/o} means ``without''. Best results are \textbf{bold}.}
\label{tab:understanding_ablation}
\centering
\resizebox{0.8\textwidth}{!}{
\begin{tabular}{l c cc cc cc cc}
\toprule
\multirow{2}{*}{\textbf{Methods}} &
\multicolumn{2}{c}{\textbf{MSVD-QA}} &
\multicolumn{2}{c}{\textbf{MSRVTT-QA}} &
\multicolumn{2}{c}{\textbf{TGIF-QA}} &
\multicolumn{2}{c}{\textbf{ActivityNet-QA}} \\
\cmidrule(lr){2-3}\cmidrule(lr){4-5}\cmidrule(lr){6-7}\cmidrule(lr){8-9}
 & Acc$\uparrow$ & Score$\uparrow$ & Acc$\uparrow$ & Score$\uparrow$ & Acc$\uparrow$ & Score$\uparrow$ & Acc$\uparrow$ & Score$\uparrow$ \\
\midrule
UniVid (Base)            & 64.1 & 3.3 & 48.9 & 2.8 & 54.2 & 3.0 & 39.8 & 3.0 \\
UniVid (w/o finetune)    & 71.1 & 3.9 & 52.2 & 3.0 & 63.5 & 3.6 & 46.5 & 3.2 \\
UniVid (w/o Reflection)   & 73.1 & 4.0 & 55.0 & 3.1 & 64.6 & 3.6 & 52.0 & 3.4 \\
UniVid (Full)           & \textbf{80.1} & \textbf{4.2} & \textbf{61.4} & \textbf{3.4} & \textbf{75.0} & \textbf{4.1} & \textbf{58.8} & \textbf{3.6} \\
\midrule
\end{tabular}
}
\vspace{-0.4cm}
\end{table}

\subsection{Ablation Study}
\label{ablation}
\paragraph{Ablation on video generation.}
Tab.~\ref{tab:video_ablation} quantifies the effect of removing key components of UniVid. Excluding multi-level language modeling (w/o MLLM) reduces semantic fidelity, especially in spatial and appearance consistency, highlighting the importance of multi-scale text grounding. Disabling Temperature Modality Alignment (w/o TMA) further degrades motion smoothness and temporal coherence. Fig.~\ref{fig:ablation_tma} visualizes these issues: without TMA, generated players exhibit unnatural fingers, distorted poses, and implausible ball trajectories, whereas the full UniVid produces coherent jump shots with realistic ball arcs. Qualitative comparisons in Fig.~\ref{fig:generation_comparasion} confirm that UniVid consistently avoids missing objects and deformations that plague prior models, achieving both semantic plausibility and temporal stability.

\paragraph{Ablation on video understanding.}
Tab.~\ref{tab:understanding_ablation} compares four variants: a lightweight base model without our training or reasoning additions, a version w/o finetune that removes Stage-II video-QA finetuning, a version w/o Reflection that keeps finetuning but disables the Pyramid Reflection loop, and the Full UniVid. Finetuning the understanding branch on ActivityNet-QA style instruction data already yields clear gains over the base, indicating that modest, task-aligned supervision substantially improves cross-modal grounding. Adding Pyramid Reflection further boosts accuracy, with similar trends in the QA scores, confirming that query-driven keyframe selection plus the Actor–Evaluator–Reflector loop improves temporal coherence and evidence retrieval. Overall, the full system combines data-efficient tuning with iterative reasoning to deliver competitive results across all four benchmarks.

\subsection{Limitation and Future Work}
Our current design couples an AR MLLM with a DiT-based video decoder, which leverages both strengths but increases compute and memory, leading to slower generation. The understanding pipeline is keyframe-driven rather than natively video-aware, so it works well on short clips but struggles with long videos and fine motion. In future work, we will reduce coupling cost and integrate native video encoders such as Video ViT and 3D VAE to handle much longer videos, albeit at the expense of substantial training.

%% file: iclr2026_conference.bbl
\begin{thebibliography}{76}
\providecommand{\natexlab}[1]{#1}
\providecommand{\url}[1]{\texttt{#1}}
\expandafter\ifx\csname urlstyle\endcsname\relax
  \providecommand{\doi}[1]{doi: #1}\else
  \providecommand{\doi}{doi: \begingroup \urlstyle{rm}\Url}\fi

\bibitem[Ainslie et~al.(2023)Ainslie, Lee{-}Thorp, de~Jong, Zemlyanskiy, Lebr{\'{o}}n, and Sanghai]{GQA}
Joshua Ainslie, James Lee{-}Thorp, Michiel de~Jong, Yury Zemlyanskiy, Federico Lebr{\'{o}}n, and Sumit Sanghai.
\newblock {GQA:} training generalized multi-query transformer models from multi-head checkpoints.
\newblock In Houda Bouamor, Juan Pino, and Kalika Bali (eds.), \emph{Proceedings of the 2023 Conference on Empirical Methods in Natural Language Processing, {EMNLP} 2023, Singapore, December 6-10, 2023}, pp.\  4895--4901. Association for Computational Linguistics, 2023.
\newblock \doi{10.18653/V1/2023.EMNLP-MAIN.298}.
\newblock URL \url{https://doi.org/10.18653/v1/2023.emnlp-main.298}.

\bibitem[Arnab et~al.(2021)Arnab, Dehghani, Heigold, Sun, Lucic, and Schmid]{vivit}
Anurag Arnab, Mostafa Dehghani, Georg Heigold, Chen Sun, Mario Lucic, and Cordelia Schmid.
\newblock Vivit: {A} video vision transformer.
\newblock In \emph{2021 {IEEE/CVF} International Conference on Computer Vision, {ICCV} 2021, Montreal, QC, Canada, October 10-17, 2021}, pp.\  6816--6826, 2021.

\bibitem[Bai et~al.(2025)Bai, Chen, Liu, Wang, Ge, Song, Dang, Wang, Wang, Tang, Zhong, Zhu, Yang, Li, Wan, Wang, Ding, Fu, Xu, Ye, Zhang, Xie, Cheng, Zhang, Yang, Xu, and Lin]{Qwen2.5-VL}
Shuai Bai, Keqin Chen, Xuejing Liu, Jialin Wang, Wenbin Ge, Sibo Song, Kai Dang, Peng Wang, Shijie Wang, Jun Tang, Humen Zhong, Yuanzhi Zhu, Ming{-}Hsuan Yang, Zhaohai Li, Jianqiang Wan, Pengfei Wang, Wei Ding, Zheren Fu, Yiheng Xu, Jiabo Ye, Xi~Zhang, Tianbao Xie, Zesen Cheng, Hang Zhang, Zhibo Yang, Haiyang Xu, and Junyang Lin.
\newblock Qwen2.5-vl technical report.
\newblock \emph{CoRR}, abs/2502.13923, 2025.

\bibitem[Bertasius et~al.(2021)Bertasius, Wang, and Torresani]{timesformer}
Gedas Bertasius, Heng Wang, and Lorenzo Torresani.
\newblock Is space-time attention all you need for video understanding?
\newblock In \emph{Proceedings of the 38th International Conference on Machine Learning, {ICML} 2021, 18-24 July 2021, Virtual Event}, pp.\  813--824, 2021.

\bibitem[Blattmann et~al.(2023{\natexlab{a}})Blattmann, Dockhorn, Kulal, Mendelevitch, Kilian, Lorenz, Levi, English, Voleti, Letts, Jampani, and Rombach]{SVD}
Andreas Blattmann, Tim Dockhorn, Sumith Kulal, Daniel Mendelevitch, Maciej Kilian, Dominik Lorenz, Yam Levi, Zion English, Vikram Voleti, Adam Letts, Varun Jampani, and Robin Rombach.
\newblock Stable video diffusion: Scaling latent video diffusion models to large datasets.
\newblock \emph{CoRR}, abs/2311.15127, 2023{\natexlab{a}}.

\bibitem[Blattmann et~al.(2023{\natexlab{b}})Blattmann, Dockhorn, Kulal, Mendelevitch, Kilian, Lorenz, Levi, English, Voleti, Letts, et~al.]{blattmann2023stable}
Andreas Blattmann, Tim Dockhorn, Sumith Kulal, Daniel Mendelevitch, Maciej Kilian, Dominik Lorenz, Yam Levi, Zion English, Vikram Voleti, Adam Letts, et~al.
\newblock Stable video diffusion: Scaling latent video diffusion models to large datasets.
\newblock \emph{arXiv preprint arXiv:2311.15127}, 2023{\natexlab{b}}.

\bibitem[Blattmann et~al.(2023{\natexlab{c}})Blattmann, Rombach, Ling, Dockhorn, Kim, Fidler, and Kreis]{blattmann2023align}
Andreas Blattmann, Robin Rombach, Huan Ling, Tim Dockhorn, Seung~Wook Kim, Sanja Fidler, and Karsten Kreis.
\newblock Align your latents: High-resolution video synthesis with latent diffusion models.
\newblock In \emph{Proceedings of the IEEE/CVF conference on computer vision and pattern recognition}, pp.\  22563--22575, 2023{\natexlab{c}}.

\bibitem[Carreira et~al.(2019)Carreira, Noland, Hillier, and Zisserman]{carreira2019short}
Joao Carreira, Eric Noland, Chloe Hillier, and Andrew Zisserman.
\newblock A short note on the kinetics-700 human action dataset.
\newblock \emph{arXiv preprint arXiv:1907.06987}, 2019.

\bibitem[Chen et~al.(2024{\natexlab{a}})Chen, Zhang, Cun, Xia, Wang, Weng, and Shan]{videocrafter}
Haoxin Chen, Yong Zhang, Xiaodong Cun, Menghan Xia, Xintao Wang, Chao Weng, and Ying Shan.
\newblock Videocrafter2: Overcoming data limitations for high-quality video diffusion models.
\newblock In \emph{{IEEE/CVF} Conference on Computer Vision and Pattern Recognition, {CVPR} 2024, Seattle, WA, USA, June 16-22, 2024}, pp.\  7310--7320, 2024{\natexlab{a}}.

\bibitem[Chen et~al.(2025)Chen, Xu, Pan, Hu, Qin, Goldstein, Huang, Zhou, Xie, Savarese, Xue, Xiong, and Xu]{BILP3-o}
Jiuhai Chen, Zhiyang Xu, Xichen Pan, Yushi Hu, Can Qin, Tom Goldstein, Lifu Huang, Tianyi Zhou, Saining Xie, Silvio Savarese, Le~Xue, Caiming Xiong, and Ran Xu.
\newblock Blip3-o: {A} family of fully open unified multimodal models-architecture, training and dataset.
\newblock \emph{CoRR}, abs/2505.09568, 2025.

\bibitem[Chen et~al.(2024{\natexlab{b}})Chen, Wu, Wang, Su, Chen, Xing, Zhong, Zhang, Zhu, Lu, Li, Luo, Lu, Qiao, and Dai]{InternVL}
Zhe Chen, Jiannan Wu, Wenhai Wang, Weijie Su, Guo Chen, Sen Xing, Muyan Zhong, Qinglong Zhang, Xizhou Zhu, Lewei Lu, Bin Li, Ping Luo, Tong Lu, Yu~Qiao, and Jifeng Dai.
\newblock Intern {VL:} scaling up vision foundation models and aligning for generic visual-linguistic tasks.
\newblock In \emph{{IEEE/CVF} Conference on Computer Vision and Pattern Recognition, {CVPR} 2024, Seattle, WA, USA, June 16-22, 2024}, pp.\  24185--24198. {IEEE}, 2024{\natexlab{b}}.

\bibitem[Cuong et~al.(2025)Cuong, Tam, Chinh, and Hanh]{fluid}
Van~Duc Cuong, Ta~Dinh Tam, Tran~Duc Chinh, and Nguyen~Thi Hanh.
\newblock Fluid: Flow-latent unified integration via token distillation for expert specialization in multimodal learning, 2025.

\bibitem[Deng et~al.(2025)Deng, Zhu, Li, Gou, Li, Wang, Zhong, Yu, Nie, Song, Guang, and Fan]{BAGEL}
Chaorui Deng, Deyao Zhu, Kunchang Li, Chenhui Gou, Feng Li, Zeyu Wang, Shu Zhong, Weihao Yu, Xiaonan Nie, Ziang Song, Shi Guang, and Haoqi Fan.
\newblock Emerging properties in unified multimodal pretraining.
\newblock \emph{CoRR}, 2025.

\bibitem[Dosovitskiy et~al.(2021)Dosovitskiy, Beyer, Kolesnikov, Weissenborn, Zhai, Unterthiner, Dehghani, Minderer, Heigold, Gelly, Uszkoreit, and Houlsby]{vit}
Alexey Dosovitskiy, Lucas Beyer, Alexander Kolesnikov, Dirk Weissenborn, Xiaohua Zhai, Thomas Unterthiner, Mostafa Dehghani, Matthias Minderer, Georg Heigold, Sylvain Gelly, Jakob Uszkoreit, and Neil Houlsby.
\newblock An image is worth 16x16 words: Transformers for image recognition at scale.
\newblock In \emph{9th International Conference on Learning Representations, {ICLR} 2021, Virtual Event, Austria, May 3-7, 2021}. OpenReview.net, 2021.
\newblock URL \url{https://openreview.net/forum?id=YicbFdNTTy}.

\bibitem[Esser et~al.(2024)Esser, Kulal, Blattmann, Entezari, M{\"{u}}ller, Saini, Levi, Lorenz, Sauer, Boesel, Podell, Dockhorn, English, and Rombach]{MM-DiT}
Patrick Esser, Sumith Kulal, Andreas Blattmann, Rahim Entezari, Jonas M{\"{u}}ller, Harry Saini, Yam Levi, Dominik Lorenz, Axel Sauer, Frederic Boesel, Dustin Podell, Tim Dockhorn, Zion English, and Robin Rombach.
\newblock Scaling rectified flow transformers for high-resolution image synthesis.
\newblock In \emph{Forty-first International Conference on Machine Learning, {ICML} 2024, Vienna, Austria, July 21-27, 2024}. OpenReview.net, 2024.
\newblock URL \url{https://openreview.net/forum?id=FPnUhsQJ5B}.

\bibitem[Fan et~al.(2021)Fan, Xiong, Mangalam, Li, Yan, Malik, and Feichtenhofer]{MViT}
Haoqi Fan, Bo~Xiong, Karttikeya Mangalam, Yanghao Li, Zhicheng Yan, Jitendra Malik, and Christoph Feichtenhofer.
\newblock Multiscale vision transformers.
\newblock In \emph{Proceedings of the IEEE/CVF international conference on computer vision}, pp.\  6824--6835, 2021.

\bibitem[Fan et~al.(2025)Fan, Si, Song, Yang, He, Zhuo, Huang, Dong, He, Pan, et~al.]{vchitect2025}
Weichen Fan, Chenyang Si, Junhao Song, Zhenyu Yang, Yinan He, Long Zhuo, Ziqi Huang, Ziyue Dong, Jingwen He, Dongwei Pan, et~al.
\newblock Vchitect-2.0: Parallel transformer for scaling up video diffusion models.
\newblock \emph{arXiv preprint arXiv:2501.08453}, 2025.

\bibitem[Fu et~al.(2024)Fu, Yeh, Zha, Wang, Li, Shaw, Li, and Chen]{easyanimate2024}
Jaskie Fu, Kun-Hao Yeh, Zhaofan Zha, Xinyu Wang, Chenghao Li, Han-Yi Shaw, Chao-Yi Li, and Pin-Yu Chen.
\newblock Easyanimate: An easy-to-use framework for creating high-quality and controllable videos from a single image.
\newblock \emph{arXiv preprint arXiv:2403.04416}, 2024.

\bibitem[Girdhar et~al.(2023)Girdhar, El-Nouby, Singh, Alwala, Joulin, and Misra]{girdhar2023omnimae}
Rohit Girdhar, Alaaeldin El-Nouby, Mannat Singh, Kalyan~Vasudev Alwala, Armand Joulin, and Ishan Misra.
\newblock Omnimae: Single model masked pretraining on images and videos.
\newblock In \emph{Proceedings of the IEEE/CVF conference on computer vision and pattern recognition}, pp.\  10406--10417, 2023.

\bibitem[Henry et~al.(2020)Henry, Dachapally, Pawar, and Chen]{QKNorm}
Alex Henry, Prudhvi~Raj Dachapally, Shubham~Shantaram Pawar, and Yuxuan Chen.
\newblock Query-key normalization for transformers.
\newblock In Trevor Cohn, Yulan He, and Yang Liu (eds.), \emph{Findings of the Association for Computational Linguistics: {EMNLP} 2020, Online Event, 16-20 November 2020}, volume {EMNLP} 2020 of \emph{Findings of {ACL}}, pp.\  4246--4253. Association for Computational Linguistics, 2020.
\newblock \doi{10.18653/V1/2020.FINDINGS-EMNLP.379}.
\newblock URL \url{https://doi.org/10.18653/v1/2020.findings-emnlp.379}.

\bibitem[Ho et~al.(2022{\natexlab{a}})Ho, Chan, Saharia, Whang, Gao, Gritsenko, Kingma, Poole, Norouzi, Fleet, and Salimans]{Imagen}
Jonathan Ho, William Chan, Chitwan Saharia, Jay Whang, Ruiqi Gao, Alexey~A. Gritsenko, Diederik~P. Kingma, Ben Poole, Mohammad Norouzi, David~J. Fleet, and Tim Salimans.
\newblock Imagen video: High definition video generation with diffusion models.
\newblock \emph{CoRR}, 2022{\natexlab{a}}.

\bibitem[Ho et~al.(2022{\natexlab{b}})Ho, Salimans, Gritsenko, Chan, Norouzi, and Fleet]{vdm}
Jonathan Ho, Tim Salimans, Alexey~A. Gritsenko, William Chan, Mohammad Norouzi, and David~J. Fleet.
\newblock Video diffusion models.
\newblock In \emph{Advances in Neural Information Processing Systems 35: Annual Conference on Neural Information Processing Systems 2022, NeurIPS 2022, New Orleans, LA, USA, November 28 - December 9, 2022}, 2022{\natexlab{b}}.

\bibitem[Huang et~al.(2024)Huang, He, Yu, Zhang, Si, Jiang, Zhang, Wu, Jin, Chanpaisit, et~al.]{huang2024vbench}
Ziqi Huang, Yinan He, Jiashuo Yu, Fan Zhang, Chenyang Si, Yuming Jiang, Yuanhan Zhang, Tianxing Wu, Qingyang Jin, Nattapol Chanpaisit, et~al.
\newblock Vbench: Comprehensive benchmark suite for video generative models.
\newblock In \emph{Proceedings of the IEEE/CVF Conference on Computer Vision and Pattern Recognition}, pp.\  21807--21818, 2024.

\bibitem[Jang et~al.(2017)Jang, Song, Yu, Kim, and Kim]{TGIF}
Yunseok Jang, Yale Song, Youngjae Yu, Youngjin Kim, and Gunhee Kim.
\newblock {TGIF-QA:} toward spatio-temporal reasoning in visual question answering.
\newblock In \emph{2017 {IEEE} Conference on Computer Vision and Pattern Recognition, {CVPR} 2017, Honolulu, HI, USA, July 21-26, 2017}, pp.\  1359--1367. {IEEE} Computer Society, 2017.

\bibitem[Jin et~al.(2024)Jin, Takanobu, Zhang, Cao, and Yuan]{Chat-Univi2024}
Peng Jin, Ryuichi Takanobu, Wancai Zhang, Xiaochun Cao, and Li~Yuan.
\newblock Chat-univi: Unified visual representation empowers large language models with image and video understanding.
\newblock In \emph{{IEEE/CVF} Conference on Computer Vision and Pattern Recognition, {CVPR} 2024, Seattle, WA, USA, June 16-22, 2024}, pp.\  13700--13710. {IEEE}, 2024.
\newblock \doi{10.1109/CVPR52733.2024.01300}.

\bibitem[Kim et~al.(2024)Kim, Choi, Lee, and Rhee]{IG-VLM}
Wonkyun Kim, Changin Choi, Wonseok Lee, and Wonjong Rhee.
\newblock An image grid can be worth a video: Zero-shot video question answering using a vlm, 2024.
\newblock URL \url{https://arxiv.org/abs/2403.18406}.

\bibitem[Kingma \& Welling(2019)Kingma and Welling]{VAE}
Diederik~P. Kingma and Max Welling.
\newblock An introduction to variational autoencoders.
\newblock \emph{Found. Trends Mach. Learn.}, 12\penalty0 (4):\penalty0 307--392, 2019.
\newblock \doi{10.1561/2200000056}.
\newblock URL \url{https://doi.org/10.1561/2200000056}.

\bibitem[Kong et~al.(2024)Kong, Tian, Zhang, Min, Dai, Zhou, Xiong, Li, Wu, Zhang, et~al.]{hunyuanvideo2024}
Weijie Kong, Qi~Tian, Zijian Zhang, Rox Min, Zuozhuo Dai, Jin Zhou, Jiangfeng Xiong, Xin Li, Bo~Wu, Jianwei Zhang, et~al.
\newblock Hunyuanvideo: A systematic framework for large video generative models.
\newblock \emph{arXiv preprint arXiv:2412.03603}, 2024.

\bibitem[Li et~al.(2023)Li, He, Wang, Li, Wang, Luo, Wang, Wang, and Qiao]{Videochat2023}
Kunchang Li, Yinan He, Yi~Wang, Yizhuo Li, Wenhai Wang, Ping Luo, Yali Wang, Limin Wang, and Yu~Qiao.
\newblock Videochat: Chat-centric video understanding.
\newblock \emph{CoRR}, abs/2305.06355, 2023.

\bibitem[Li et~al.(2024)Li, Wang, Cai, Qi, Wang, Zhang, Song, Jiang, Huang, and Wang]{li2024unifiedmllm}
Zhaowei Li, Wei Wang, YiQing Cai, Xu~Qi, Pengyu Wang, Dong Zhang, Hang Song, Botian Jiang, Zhida Huang, and Tao Wang.
\newblock Unifiedmllm: Enabling unified representation for multi-modal multi-tasks with large language model.
\newblock \emph{arXiv preprint arXiv:2408.02503}, 2024.

\bibitem[Lin et~al.(2024)Lin, Ye, Zhu, Cui, Ning, Jin, and Yuan]{Video-LLava}
Bin Lin, Yang Ye, Bin Zhu, Jiaxi Cui, Munan Ning, Peng Jin, and Li~Yuan.
\newblock Video-llava: Learning united visual representation by alignment before projection.
\newblock In Yaser Al{-}Onaizan, Mohit Bansal, and Yun{-}Nung Chen (eds.), \emph{Proceedings of the 2024 Conference on Empirical Methods in Natural Language Processing, {EMNLP} 2024, Miami, FL, USA, November 12-16, 2024}, pp.\  5971--5984. Association for Computational Linguistics, 2024.

\bibitem[Liu et~al.(2025)Liu, Zhang, Li, Bai, Han, Tang, Xing, Wu, Yang, Chen, et~al.]{liu2025fpsattention}
Akide Liu, Zeyu Zhang, Zhexin Li, Xuehai Bai, Yizeng Han, Jiasheng Tang, Yuanjie Xing, Jichao Wu, Mingyang Yang, Weihua Chen, et~al.
\newblock Fpsattention: Training-aware fp8 and sparsity co-design for fast video diffusion.
\newblock \emph{arXiv preprint arXiv:2506.04648}, 2025.

\bibitem[Liu et~al.(2024)Liu, Li, Ge, Li, Shan, and Li]{BT-Adapter2024}
Ruyang Liu, Chen Li, Yixiao Ge, Thomas~H. Li, Ying Shan, and Ge~Li.
\newblock Bt-adapter: Video conversation is feasible without video instruction tuning.
\newblock In \emph{{IEEE/CVF} Conference on Computer Vision and Pattern Recognition, {CVPR} 2024, Seattle, WA, USA, June 16-22, 2024}, pp.\  13658--13667. {IEEE}, 2024.
\newblock \doi{10.1109/CVPR52733.2024.01296}.

\bibitem[Liu et~al.(2022)Liu, Ning, Cao, Wei, Zhang, Lin, and Hu]{liu2022video}
Ze~Liu, Jia Ning, Yue Cao, Yixuan Wei, Zheng Zhang, Stephen Lin, and Han Hu.
\newblock Video swin transformer.
\newblock In \emph{Proceedings of the IEEE/CVF conference on computer vision and pattern recognition}, pp.\  3202--3211, 2022.

\bibitem[Lu et~al.(2023)Lu, Peng, Cheng, Galley, Chang, Wu, Zhu, and Gao]{Chameleon}
Pan Lu, Baolin Peng, Hao Cheng, Michel Galley, Kai{-}Wei Chang, Ying~Nian Wu, Song{-}Chun Zhu, and Jianfeng Gao.
\newblock Chameleon: Plug-and-play compositional reasoning with large language models.
\newblock In Alice Oh, Tristan Naumann, Amir Globerson, Kate Saenko, Moritz Hardt, and Sergey Levine (eds.), \emph{Advances in Neural Information Processing Systems 36: Annual Conference on Neural Information Processing Systems 2023, NeurIPS 2023, New Orleans, LA, USA, December 10 - 16, 2023}, 2023.

\bibitem[Luo et~al.(2023)Luo, Zhao, Yang, Dong, Qiu, Lu, Wang, and Wei]{Valley-v3}
Ruipu Luo, Ziwang Zhao, Min Yang, Junwei Dong, Minghui Qiu, Pengcheng Lu, Tao Wang, and Zhongyu Wei.
\newblock Valley: Video assistant with large language model enhanced ability.
\newblock \emph{CoRR}, abs/2306.07207, 2023.

\bibitem[Lv et~al.(2025)Lv, Pan, Si, Chen, Zuo, Liu, and Wong]{lv2025rethinkingcrossmodalinteractionmultimodal}
Zhengyao Lv, Tianlin Pan, Chenyang Si, Zhaoxi Chen, Wangmeng Zuo, Ziwei Liu, and Kwan-Yee~K. Wong.
\newblock Rethinking cross-modal interaction in multimodal diffusion transformers, 2025.
\newblock URL \url{https://arxiv.org/abs/2506.07986}.

\bibitem[Maaz et~al.(2024)Maaz, Rasheed, Khan, and Khan]{VideoChatGPT2024}
Muhammad Maaz, Hanoona~Abdul Rasheed, Salman Khan, and Fahad Khan.
\newblock Video-chatgpt: Towards detailed video understanding via large vision and language models.
\newblock In Lun{-}Wei Ku, Andre Martins, and Vivek Srikumar (eds.), \emph{Proceedings of the 62nd Annual Meeting of the Association for Computational Linguistics (Volume 1: Long Papers), {ACL} 2024, Bangkok, Thailand, August 11-16, 2024}, pp.\  12585--12602. Association for Computational Linguistics, 2024.
\newblock \doi{10.18653/V1/2024.ACL-LONG.679}.

\bibitem[Melnik et~al.(2024)Melnik, Ljubljanac, Lu, Yan, Ren, and Ritter]{vdm_survey}
Andrew Melnik, Michal Ljubljanac, Cong Lu, Qi~Yan, Weiming Ren, and Helge~J. Ritter.
\newblock Video diffusion models: {A} survey.
\newblock \emph{Trans. Mach. Learn. Res.}, 2024.

\bibitem[Meng et~al.(2024)Meng, Yang, Tian, Dai, Wu, Gao, and Jiang]{DeepStack-L}
Lingchen Meng, Jianwei Yang, Rui Tian, Xiyang Dai, Zuxuan Wu, Jianfeng Gao, and Yu{-}Gang Jiang.
\newblock Deepstack: Deeply stacking visual tokens is surprisingly simple and effective for lmms.
\newblock In Amir Globersons, Lester Mackey, Danielle Belgrave, Angela Fan, Ulrich Paquet, Jakub~M. Tomczak, and Cheng Zhang (eds.), \emph{Advances in Neural Information Processing Systems 38: Annual Conference on Neural Information Processing Systems 2024, NeurIPS 2024, Vancouver, BC, Canada, December 10 - 15, 2024}, 2024.
\newblock URL \url{http://papers.nips.cc/paper\_files/paper/2024/hash/29cd7f8331d13ede6dc6d6ef3dfacb70-Abstract-Conference.html}.

\bibitem[MiniMax(2024)]{minimax2024video}
MiniMax.
\newblock Minimax video generation api is now available.
\newblock \url{https://www.minimaxi.com/en/news/video-generation-api}, October 2024.
\newblock Accessed: 2025-07-24.

\bibitem[Pan et~al.(2021)Pan, Song, Yang, Jiang, and Liu]{pan2021videomoco}
Tian Pan, Yibing Song, Tianyu Yang, Wenhao Jiang, and Wei Liu.
\newblock Videomoco: Contrastive video representation learning with temporally adversarial examples.
\newblock In \emph{Proceedings of the IEEE/CVF conference on computer vision and pattern recognition}, pp.\  11205--11214, 2021.

\bibitem[Peebles \& Xie(2023)Peebles and Xie]{DBLP:conf/iccv/PeeblesX23}
William Peebles and Saining Xie.
\newblock Scalable diffusion models with transformers.
\newblock In \emph{{IEEE/CVF} International Conference on Computer Vision, {ICCV} 2023, Paris, France, October 1-6, 2023}, pp.\  4172--4182. {IEEE}, 2023.
\newblock \doi{10.1109/ICCV51070.2023.00387}.
\newblock URL \url{https://doi.org/10.1109/ICCV51070.2023.00387}.

\bibitem[Piergiovanni et~al.(2022)Piergiovanni, Morton, Kuo, Ryoo, and Angelova]{msrvtt}
A.~J. Piergiovanni, Kairo Morton, Weicheng Kuo, Michael~S. Ryoo, and Anelia Angelova.
\newblock Video question answering with iterative video-text co-tokenization.
\newblock In Shai Avidan, Gabriel~J. Brostow, Moustapha Ciss{\'{e}}, Giovanni~Maria Farinella, and Tal Hassner (eds.), \emph{Computer Vision - {ECCV} 2022 - 17th European Conference, Tel Aviv, Israel, October 23-27, 2022, Proceedings, Part {XXXVI}}, volume 13696 of \emph{Lecture Notes in Computer Science}, pp.\  76--94. Springer, 2022.

\bibitem[Podell et~al.(2024)Podell, English, Lacey, Blattmann, Dockhorn, M{\"{u}}ller, Penna, and Rombach]{SDXL}
Dustin Podell, Zion English, Kyle Lacey, Andreas Blattmann, Tim Dockhorn, Jonas M{\"{u}}ller, Joe Penna, and Robin Rombach.
\newblock {SDXL:} improving latent diffusion models for high-resolution image synthesis.
\newblock In \emph{The Twelfth International Conference on Learning Representations, {ICLR} 2024, Vienna, Austria, May 7-11, 2024}. OpenReview.net, 2024.

\bibitem[Radford et~al.(2021)Radford, Kim, Hallacy, Ramesh, Goh, Agarwal, Sastry, Askell, Mishkin, Clark, et~al.]{radford2021learning}
Alec Radford, Jong~Wook Kim, Chris Hallacy, Aditya Ramesh, Gabriel Goh, Sandhini Agarwal, Girish Sastry, Amanda Askell, Pamela Mishkin, Jack Clark, et~al.
\newblock Learning transferable visual models from natural language supervision.
\newblock In \emph{International conference on machine learning}, pp.\  8748--8763. PmLR, 2021.

\bibitem[Runway(2024)]{runway2024gen3}
Runway.
\newblock Gen-3 alpha: A new frontier for video generation.
\newblock Technical report, Runway, July 2024.
\newblock Accessed: 2025-07-24.

\bibitem[Shazeer(2020)]{GLU}
Noam Shazeer.
\newblock {GLU} variants improve transformer.
\newblock \emph{CoRR}, abs/2002.05202, 2020.
\newblock URL \url{https://arxiv.org/abs/2002.05202}.

\bibitem[Shi et~al.(2025)Shi, Zhang, Wu, Liang, Fang, Chen, and Zhao]{shi2025presentagent}
Jingwei Shi, Zeyu Zhang, Biao Wu, Yanjie Liang, Meng Fang, Ling Chen, and Yang Zhao.
\newblock Presentagent: Multimodal agent for presentation video generation.
\newblock \emph{arXiv preprint arXiv:2507.04036}, 2025.

\bibitem[Shi et~al.(2020)Shi, Zhou, Qiu, and Zhu]{shi2020improvingimagecaptioningbetter}
Zhan Shi, Xu~Zhou, Xipeng Qiu, and Xiaodan Zhu.
\newblock Improving image captioning with better use of captions, 2020.
\newblock URL \url{https://arxiv.org/abs/2006.11807}.

\bibitem[Shinn et~al.(2023)Shinn, Cassano, Gopinath, Narasimhan, and Yao]{reflexion}
Noah Shinn, Federico Cassano, Ashwin Gopinath, Karthik Narasimhan, and Shunyu Yao.
\newblock Reflexion: language agents with verbal reinforcement learning.
\newblock In Alice Oh, Tristan Naumann, Amir Globerson, Kate Saenko, Moritz Hardt, and Sergey Levine (eds.), \emph{Advances in Neural Information Processing Systems 36: Annual Conference on Neural Information Processing Systems 2023, NeurIPS 2023, New Orleans, LA, USA, December 10 - 16, 2023}, 2023.
\newblock URL \url{http://papers.nips.cc/paper\_files/paper/2023/hash/1b44b878bb782e6954cd888628510e90-Abstract-Conference.html}.

\bibitem[Skorokhodov et~al.(2022)Skorokhodov, Tulyakov, and Elhoseiny]{stylegan-v}
Ivan Skorokhodov, Sergey Tulyakov, and Mohamed Elhoseiny.
\newblock Stylegan-v: {A} continuous video generator with the price, image quality and perks of stylegan2.
\newblock In \emph{{IEEE/CVF} Conference on Computer Vision and Pattern Recognition, {CVPR} 2022, New Orleans, LA, USA, June 18-24, 2022}, pp.\  3616--3626, 2022.

\bibitem[Su et~al.(2024)Su, Ahmed, Lu, Pan, Bo, and Liu]{RoPE}
Jianlin Su, Murtadha H.~M. Ahmed, Yu~Lu, Shengfeng Pan, Wen Bo, and Yunfeng Liu.
\newblock Roformer: Enhanced transformer with rotary position embedding.
\newblock \emph{Neurocomputing}, 568:\penalty0 127063, 2024.
\newblock \doi{10.1016/J.NEUCOM.2023.127063}.
\newblock URL \url{https://doi.org/10.1016/j.neucom.2023.127063}.

\bibitem[Tan et~al.(2025)Tan, Yang, Qin, Gong, Yang, and Li]{Omni-Video}
Zhiyu Tan, Hao Yang, Luozheng Qin, Jia Gong, Mengping Yang, and Hao Li.
\newblock Omni-video: Democratizing unified video understanding and generation.
\newblock \emph{CoRR}, abs/2507.06119, 2025.

\bibitem[Technology(2025)]{kling2025}
Kuaishou Technology.
\newblock Kling.
\newblock \url{https://klingai.kuaishou.com/}, 2025.
\newblock Accessed: 2025-07-24.

\bibitem[Tong et~al.(2022)Tong, Song, Wang, and Wang]{videomae}
Zhan Tong, Yibing Song, Jue Wang, and Limin Wang.
\newblock Videomae: Masked autoencoders are data-efficient learners for self-supervised video pre-training.
\newblock In \emph{Advances in Neural Information Processing Systems 35: Annual Conference on Neural Information Processing Systems 2022, NeurIPS 2022, New Orleans, LA, USA, November 28 - December 9, 2022}, 2022.

\bibitem[Tschannen et~al.(2025)Tschannen, Gritsenko, Wang, Naeem, Alabdulmohsin, Parthasarathy, Evans, Beyer, Xia, Mustafa, H{\'{e}}naff, Harmsen, Steiner, and Zhai]{SigLip2}
Michael Tschannen, Alexey~A. Gritsenko, Xiao Wang, Muhammad~Ferjad Naeem, Ibrahim Alabdulmohsin, Nikhil Parthasarathy, Talfan Evans, Lucas Beyer, Ye~Xia, Basil Mustafa, Olivier~J. H{\'{e}}naff, Jeremiah Harmsen, Andreas Steiner, and Xiaohua Zhai.
\newblock Siglip 2: Multilingual vision-language encoders with improved semantic understanding, localization, and dense features.
\newblock \emph{CoRR}, abs/2502.14786, 2025.

\bibitem[Tulyakov et~al.(2018)Tulyakov, Liu, Yang, and Kautz]{MoCoGAN}
Sergey Tulyakov, Ming{-}Yu Liu, Xiaodong Yang, and Jan Kautz.
\newblock Mocogan: Decomposing motion and content for video generation.
\newblock In \emph{2018 {IEEE} Conference on Computer Vision and Pattern Recognition, {CVPR} 2018, Salt Lake City, UT, USA, June 18-22, 2018}, pp.\  1526--1535, 2018.

\bibitem[Wang et~al.(2025)Wang, Ai, Wen, Mao, Xie, Chen, Yu, Zhao, Yang, Zeng, Wang, Zhang, Zhou, Wang, Chen, Zhu, Zhao, Yan, Huang, Meng, Zhang, Li, Wu, Chu, Feng, Zhang, Sun, Fang, Wang, Gui, Weng, Shen, Lin, Wang, Wang, Zhou, Wang, Shen, Yu, Shi, Huang, Xu, Kou, Lv, Li, Liu, Wang, Zhang, Huang, Li, Wu, Liu, Pan, Zheng, Hong, Shi, Feng, Jiang, Han, Wu, and Liu]{Wan}
Ang Wang, Baole Ai, Bin Wen, Chaojie Mao, Chen{-}Wei Xie, Di~Chen, Feiwu Yu, Haiming Zhao, Jianxiao Yang, Jianyuan Zeng, Jiayu Wang, Jingfeng Zhang, Jingren Zhou, Jinkai Wang, Jixuan Chen, Kai Zhu, Kang Zhao, Keyu Yan, Lianghua Huang, Xiaofeng Meng, Ningyi Zhang, Pandeng Li, Pingyu Wu, Ruihang Chu, Ruili Feng, Shiwei Zhang, Siyang Sun, Tao Fang, Tianxing Wang, Tianyi Gui, Tingyu Weng, Tong Shen, Wei Lin, Wei Wang, Wei Wang, Wenmeng Zhou, Wente Wang, Wenting Shen, Wenyuan Yu, Xianzhong Shi, Xiaoming Huang, Xin Xu, Yan Kou, Yangyu Lv, Yifei Li, Yijing Liu, Yiming Wang, Yingya Zhang, Yitong Huang, Yong Li, You Wu, Yu~Liu, Yulin Pan, Yun Zheng, Yuntao Hong, Yupeng Shi, Yutong Feng, Zeyinzi Jiang, Zhen Han, Zhi{-}Fan Wu, and Ziyu Liu.
\newblock Wan: Open and advanced large-scale video generative models.
\newblock \emph{CoRR}, abs/2503.20314, 2025.

\bibitem[Wang et~al.(2024{\natexlab{a}})Wang, Bai, Tan, Wang, Fan, Bai, Chen, Liu, Wang, Ge, Fan, Dang, Du, Ren, Men, Liu, Zhou, Zhou, and Lin]{Qwen2-VL}
Peng Wang, Shuai Bai, Sinan Tan, Shijie Wang, Zhihao Fan, Jinze Bai, Keqin Chen, Xuejing Liu, Jialin Wang, Wenbin Ge, Yang Fan, Kai Dang, Mengfei Du, Xuancheng Ren, Rui Men, Dayiheng Liu, Chang Zhou, Jingren Zhou, and Junyang Lin.
\newblock Qwen2-vl: Enhancing vision-language model's perception of the world at any resolution.
\newblock \emph{CoRR}, abs/2409.12191, 2024{\natexlab{a}}.

\bibitem[Wang et~al.(2023)Wang, Bao, Dong, Bjorck, Peng, Liu, Aggarwal, Mohammed, Singhal, Som, et~al.]{wang2023image}
Wenhui Wang, Hangbo Bao, Li~Dong, Johan Bjorck, Zhiliang Peng, Qiang Liu, Kriti Aggarwal, Owais~Khan Mohammed, Saksham Singhal, Subhojit Som, et~al.
\newblock Image as a foreign language: Beit pretraining for vision and vision-language tasks.
\newblock In \emph{Proceedings of the IEEE/CVF Conference on Computer Vision and Pattern Recognition}, pp.\  19175--19186, 2023.

\bibitem[Wang et~al.(2024{\natexlab{b}})Wang, Zhang, Luo, Sun, Cui, Wang, Zhang, Wang, Li, Yu, Zhao, Ao, Min, Li, Wu, Zhao, Zhang, Wang, Liu, He, Yang, Liu, Lin, Huang, and Wang]{Emu3}
Xinlong Wang, Xiaosong Zhang, Zhengxiong Luo, Quan Sun, Yufeng Cui, Jinsheng Wang, Fan Zhang, Yueze Wang, Zhen Li, Qiying Yu, Yingli Zhao, Yulong Ao, Xuebin Min, Tao Li, Boya Wu, Bo~Zhao, Bowen Zhang, Liangdong Wang, Guang Liu, Zheqi He, Xi~Yang, Jingjing Liu, Yonghua Lin, Tiejun Huang, and Zhongyuan Wang.
\newblock Emu3: Next-token prediction is all you need.
\newblock \emph{CoRR}, abs/2409.18869, 2024{\natexlab{b}}.

\bibitem[Wei et~al.(2022)Wei, Fan, Xie, Wu, Yuille, and Feichtenhofer]{maskfeat}
Chen Wei, Haoqi Fan, Saining Xie, Chao{-}Yuan Wu, Alan~L. Yuille, and Christoph Feichtenhofer.
\newblock Masked feature prediction for self-supervised visual pre-training.
\newblock In \emph{{IEEE/CVF} Conference on Computer Vision and Pattern Recognition, {CVPR} 2022, New Orleans, LA, USA, June 18-24, 2022}, pp.\  14648--14658, 2022.

\bibitem[Wu(2024)]{FreeVA}
Wenhao Wu.
\newblock Freeva: Offline {MLLM} as training-free video assistant.
\newblock \emph{CoRR}, abs/2405.07798, 2024.
\newblock \doi{10.48550/ARXIV.2405.07798}.
\newblock URL \url{https://doi.org/10.48550/arXiv.2405.07798}.

\bibitem[Xie et~al.(2025)Xie, Mao, Bai, Zhang, Wang, Lin, Gu, Chen, Yang, and Shou]{showo}
Jinheng Xie, Weijia Mao, Zechen Bai, David~Junhao Zhang, Weihao Wang, Kevin~Qinghong Lin, Yuchao Gu, Zhijie Chen, Zhenheng Yang, and Mike~Zheng Shou.
\newblock Show-o: One single transformer to unify multimodal understanding and generation.
\newblock In \emph{The Thirteenth International Conference on Learning Representations, {ICLR} 2025, Singapore, April 24-28, 2025}, 2025.

\bibitem[Xu et~al.(2024)Xu, Gao, Gan, Chen, Lai, Gang, Kang, and Dehghan]{SF-LLaVA-7B}
Mingze Xu, Mingfei Gao, Zhe Gan, Hong-You Chen, Zhengfeng Lai, Haiming Gang, Kai Kang, and Afshin Dehghan.
\newblock Slowfast-llava: A strong training-free baseline for video large language models, 2024.
\newblock URL \url{https://arxiv.org/abs/2407.15841}.

\bibitem[Yang et~al.(2024)Yang, Yang, Hui, Zheng, Yu, Zhou, Li, Li, Liu, Huang, Dong, Wei, Lin, Tang, Wang, Yang, Tu, Zhang, Ma, Yang, Xu, Zhou, Bai, He, Lin, Dang, Lu, Chen, Yang, Li, Xue, Ni, Zhang, Wang, Peng, Men, Gao, Lin, Wang, Bai, Tan, Zhu, Li, Liu, Ge, Deng, Zhou, Ren, Zhang, Wei, Ren, Liu, Fan, Yao, Zhang, Wan, Chu, Liu, Cui, Zhang, Guo, and Fan]{Qwen2}
An~Yang, Baosong Yang, Binyuan Hui, Bo~Zheng, Bowen Yu, Chang Zhou, Chengpeng Li, Chengyuan Li, Dayiheng Liu, Fei Huang, Guanting Dong, Haoran Wei, Huan Lin, Jialong Tang, Jialin Wang, Jian Yang, Jianhong Tu, Jianwei Zhang, Jianxin Ma, Jianxin Yang, Jin Xu, Jingren Zhou, Jinze Bai, Jinzheng He, Junyang Lin, Kai Dang, Keming Lu, Keqin Chen, Kexin Yang, Mei Li, Mingfeng Xue, Na~Ni, Pei Zhang, Peng Wang, Ru~Peng, Rui Men, Ruize Gao, Runji Lin, Shijie Wang, Shuai Bai, Sinan Tan, Tianhang Zhu, Tianhao Li, Tianyu Liu, Wenbin Ge, Xiaodong Deng, Xiaohuan Zhou, Xingzhang Ren, Xinyu Zhang, Xipin Wei, Xuancheng Ren, Xuejing Liu, Yang Fan, Yang Yao, Yichang Zhang, Yu~Wan, Yunfei Chu, Yuqiong Liu, Zeyu Cui, Zhenru Zhang, Zhifang Guo, and Zhihao Fan.
\newblock Qwen2 technical report.
\newblock \emph{CoRR}, abs/2407.10671, 2024.
\newblock \doi{10.48550/ARXIV.2407.10671}.
\newblock URL \url{https://doi.org/10.48550/arXiv.2407.10671}.

\bibitem[Yang et~al.(2022)Yang, Miech, Sivic, Laptev, and Schmid]{FrozenBiLM2022}
Antoine Yang, Antoine Miech, Josef Sivic, Ivan Laptev, and Cordelia Schmid.
\newblock Zero-shot video question answering via frozen bidirectional language models.
\newblock In Sanmi Koyejo, S.~Mohamed, A.~Agarwal, Danielle Belgrave, K.~Cho, and A.~Oh (eds.), \emph{Advances in Neural Information Processing Systems 35: Annual Conference on Neural Information Processing Systems 2022, NeurIPS 2022, New Orleans, LA, USA, November 28 - December 9, 2022}, 2022.

\bibitem[Yin et~al.(2025)Yin, Chen, Bai, Jiang, Li, Li, Liu, Xiang, Yu, and Zhang]{consistency_survey}
Zhiyu Yin, Kehai Chen, Xuefeng Bai, Ruili Jiang, Juntao Li, Hongdong Li, Jin Liu, Yang Xiang, Jun Yu, and Min Zhang.
\newblock Asurvey: Spatiotemporal consistency in video generation, 2025.

\bibitem[Yu et~al.(2018)Yu, Xu, Yu, Yu, Zhao, Zhuang, and Tao]{ActivityNet-QA}
Zhou Yu, Dejing Xu, Jun Yu, Ting Yu, Zhou Zhao, Yueting Zhuang, and Dacheng Tao.
\newblock Activitynet-qa: {A} dataset for understanding complex web videos via question answering.
\newblock In \emph{The Thirty-Third {AAAI} Conference on Artificial Intelligence, {AAAI} 2019, The Thirty-First Innovative Applications of Artificial Intelligence Conference, {IAAI} 2019, The Ninth {AAAI} Symposium on Educational Advances in Artificial Intelligence, {EAAI} 2019, Honolulu, Hawaii, USA, January 27 - February 1, 2019}, pp.\  9127--9134. {AAAI} Press, 2018.

\bibitem[Yuan et~al.(2024)Yuan, Chen, Zhao, yi~Wang, Zhang, Wang, Zhang, Zhao, Li, Wu, Ding, and Tang]{cogvideox2024}
Zhen Yuan, Yifei Chen, Shuo Zhao, Wen yi~Wang, Ming-Hao Zhang, Zhiping Wang, Le~Zhang, Boxi Zhao, Jian Li, Zhi-Yuan Wu, Ming Ding, and Jie Tang.
\newblock Cogvideox: A general-purpose video generation model.
\newblock \emph{arXiv preprint arXiv:2406.06511}, 2024.

\bibitem[Zhang \& Sennrich(2019)Zhang and Sennrich]{RMSNorm}
Biao Zhang and Rico Sennrich.
\newblock Root mean square layer normalization.
\newblock In Hanna~M. Wallach, Hugo Larochelle, Alina Beygelzimer, Florence d'Alch{\'{e}}{-}Buc, Emily~B. Fox, and Roman Garnett (eds.), \emph{Advances in Neural Information Processing Systems 32: Annual Conference on Neural Information Processing Systems 2019, NeurIPS 2019, December 8-14, 2019, Vancouver, BC, Canada}, pp.\  12360--12371, 2019.
\newblock URL \url{https://proceedings.neurips.cc/paper/2019/hash/1e8a19426224ca89e83cef47f1e7f53b-Abstract.html}.

\bibitem[Zhang et~al.(2023{\natexlab{a}})Zhang, Li, and Bing]{Video-LLAMA2023}
Hang Zhang, Xin Li, and Lidong Bing.
\newblock Video-llama: An instruction-tuned audio-visual language model for video understanding.
\newblock In Yansong Feng and Els Lefever (eds.), \emph{Proceedings of the 2023 Conference on Empirical Methods in Natural Language Processing, {EMNLP} 2023 - System Demonstrations, Singapore, December 6-10, 2023}, pp.\  543--553. Association for Computational Linguistics, 2023{\natexlab{a}}.

\bibitem[Zhang et~al.(2023{\natexlab{b}})Zhang, Han, Liu, Zhou, Lu, Qiao, Li, and Gao]{LLaMA-Adapter2024}
Renrui Zhang, Jiaming Han, Chris Liu, Aojun Zhou, Pan Lu, Yu~Qiao, Hongsheng Li, and Peng Gao.
\newblock Llama-adapter: Efficient fine-tuning of large language models with zero-initialized attention.
\newblock In \emph{The Twelfth International Conference on Learning Representations, {ICLR} 2024, Vienna, Austria, May 7-11, 2024}. OpenReview.net, 2023{\natexlab{b}}.

\bibitem[Zhao et~al.(2024)Zhao, Zhang, Cun, Yang, Niu, Li, Hu, and Shan]{3DVAE}
Sijie Zhao, Yong Zhang, Xiaodong Cun, Shaoshu Yang, Muyao Niu, Xiaoyu Li, Wenbo Hu, and Ying Shan.
\newblock {CV-VAE:} {A} compatible video {VAE} for latent generative video models.
\newblock In Amir Globersons, Lester Mackey, Danielle Belgrave, Angela Fan, Ulrich Paquet, Jakub~M. Tomczak, and Cheng Zhang (eds.), \emph{Advances in Neural Information Processing Systems 38: Annual Conference on Neural Information Processing Systems 2024, NeurIPS 2024, Vancouver, BC, Canada, December 10 - 15, 2024}, 2024.
\newblock URL \url{http://papers.nips.cc/paper\_files/paper/2024/hash/1787533e171dcc8549cc2eb5a4840eec-Abstract-Conference.html}.

\bibitem[Zhou et~al.(2024)Zhou, Yu, Babu, Tirumala, Yasunaga, Shamis, Kahn, Ma, Zettlemoyer, and Levy]{transfusion}
Chunting Zhou, Lili Yu, Arun Babu, Kushal Tirumala, Michihiro Yasunaga, Leonid Shamis, Jacob Kahn, Xuezhe Ma, Luke Zettlemoyer, and Omer Levy.
\newblock Transfusion: Predict the next token and diffuse images with one multi-modal model, 2024.
\newblock URL \url{https://arxiv.org/abs/2408.11039}.

\end{thebibliography}
